\documentclass[lettersize,journal]{IEEEtran}
\usepackage{cite}
\usepackage{moreverb,url}
\usepackage{booktabs}
\usepackage{siunitx}
\usepackage{multirow}
\usepackage{pifont}
\usepackage{amssymb}
\usepackage{adjustbox}
\usepackage{amsmath}
\usepackage{subfigure}
\usepackage{tabularx}
\usepackage{comment}
\usepackage{mathtools} 
\usepackage{stmaryrd}
\usepackage{tabularx} 
\usepackage{color}
\usepackage{colortbl}
\usepackage[dvipsnames]{xcolor}
\usepackage{caption}
\usepackage[dvipsnames]{xcolor}
\usepackage{pifont}       
\usepackage{bbding}
\usepackage{booktabs}
\usepackage[colorlinks,bookmarksopen,bookmarksnumbered,citecolor=red,urlcolor=red]{hyperref}
\usepackage{colortbl} 
\usepackage{arydshln}
\usepackage[dvipsnames]{xcolor}
\usepackage{pifont} 
\usepackage{tikz} 
\usepackage{dblfloatfix} 
\usepackage{graphicx}

\newcommand\BibTeX{{\rmfamily B\kern-.05em \textsc{i\kern-.025em b}\kern-.08em
T\kern-.1667em\lower.7ex\hbox{E}\kern-.125emX}}

\definecolor{C1}{HTML}{ECA8A9}
\definecolor{C2}{HTML}{74AED4}
\definecolor{C3}{HTML}{D3E2B7}
\definecolor{C4}{HTML}{CFAFD4}
\definecolor{C5}{HTML}{F7C97E}
\definecolor{C6}{HTML}{E8D5C4}
\definecolor{C7}{HTML}{EEEEEE}
\definecolor{C8}{HTML}{BCEE68}

\newcommand{\cmark}{\ding{51}}%
\newcommand{\xmark}{\ding{55}}%

\definecolor{tabfirst}{rgb}{1, 0.75, 0.7}
\definecolor{tabsecond}{rgb}{1, 0.85, 0.65}
\definecolor{tabthird}{rgb}{1, 0.96, 0.7}
\definecolor{cvprblue}{rgb}{0.21,0.49,0.74}

\begin{document}

\title{Adaptive World Memory 3D Foundation Model for Scalable 3D Mapping, Localization, and Rendering}

\author{Tianchen Deng, Guole Shen, Yilin Shen, Wenhua Wu, Yilin Fang, Ziqi Ma, Tianjun Zhang, Shenghai Yuan,  Wolfram Burgard~\IEEEmembership{Fellow,~IEEE}, Hesheng Wang~\IEEEmembership{Senior Member,~IEEE}}

\maketitle


%

    

\makeatletter{\renewcommand*{\@makefnmark}{}
 \footnotetext{Tianchen Deng, Guole Shen, Yilin Shen, Yilin Fang, Wenhua Wu, Tianjun Zhang, Hesheng Wang are with School of Automation and Intelligent Sensing, Shanghai Jiao Tong university and State Key Laboratory of Avionics Integration and Aviation System-of-Systems Synthesis, Shanghai Key Laboratory of Navigation and Location Based Services, Shanghai 200240, China. Shenghai Yuan is with Nanyang Technological University, Singapore. Wolfram Burgard is with the University
of Technology Nuremberg, Germany. This work was supported by National Key R\&D Program of China (Grant No.2024YFB4708900). It was also supported in part by the Natural Science Foundation of
China under Grant 62225309,U24A20278, 62361166632. The first three authors contribute equal to this paper. (*corresponding author: wanghesheng@sjtu.edu.cn)}
}\makeatother


\IEEEpubidadjcol

\begin{abstract}
Recent 3D foundation models enable generalizable geometric reasoning from RGB images but remain limited in persistent memory, scalability, and renderable scene modeling. We present a memory-centric 3D foundation model for scalable robotic localization, reconstruction, and Gaussian rendering. Its core is an adaptive world memory mechanism that combines transformer-based gated updates with test-time temporal-spatial regulation. Learned gates control recurrent memory propagation, while temporal state evolution and spatial observation-state consistency regulate token-wise updates and forgetting over long image sequences. To support large-scale mapping, we organize memory into local submaps and integrate progressive mapping and tracking, loop closure, and SL(4)-based global refinement to maintain local accuracy and global consistency. A Gaussian reconstruction head decodes memory-enhanced features into renderable primitives, unifying camera pose estimation, dense point-cloud reconstruction, and photorealistic rendering within a single model. Experiments on public benchmarks and self-collected datasets from diverse robotic platforms demonstrate improved trajectory accuracy, reconstruction completeness, and rendering quality over existing 3D foundation reconstruction and SLAM baselines. These results support adaptive memory as a foundation for persistent robotic world modeling. The dataset and code will be made publicly available at \href{https://github.com/dtc111111/AWM-3DFM}{https://github.com/dtc111111/AWM-3DFM}.

\end{abstract}

\section{Introduction}
\label{sec:intro}
Recent advances in 3D foundation models have substantially changed the landscape of visual 3D geometric perception. Instead of relying on hand-crafted feature matching, such as ORB-SLAM2,3~\cite{orbslam2,orbslam3}, calibrated multi-view geometry, or scene-specific optimization, models such as DUSt3R~\cite{dust3r}, MASt3R~\cite{mast3r}, VGGT~\cite{vggt}, Pi3~\cite{wang2025pi}, and their streaming variants are able to infer dense pointmaps, camera poses, and scene geometry directly from RGB observations. This paradigm provides a promising foundation for robotic 3D perception~\cite{zhang2025zisvfm,talak2023certifiable,shen2026unilgl} and scene representation~\cite{deng2025best3dscenerepresentation}, where a robot is expected to localize itself, reconstruct previously unseen environments, and maintain a consistent representation of the surrounding world during online operation. However, despite their impressive generalization ability, current 3D foundation models are still far from being complete robotic world models. Most of them are designed for pairwise, short-window, or weakly stateful inference, and therefore lack a persistent, adaptive, and globally consistent memory that can support long-term localization, reconstruction, and rendering.

A central limitation of existing 3D foundation models lies in how they handle temporal information during long-sequence inference. Feed-forward models typically process input views jointly or within short temporal windows, which leads to rapidly increasing memory consumption and limited scalability as the sequence length grows. Recent streaming methods attempt to address this issue by reducing the cost of global temporal reasoning. For example, StreamVGGT~\cite{zhuo2025streaming}, InfiniteVGGT~\cite{yuan2026infinitevggt} sparsifie global attention, thereby improving the efficiency of long-sequence encoding. Other recurrent reconstruction models explicitly maintain memory across frames: Spann3R~\cite{spann3r} utilizes an external spatial memory for incremental reconstruction, while CUT3R~\cite{cut3r} incorporates recurrent states for sequential integration. However, the state updates in these methods are often passive, uniform, or weakly adaptive. As a result, low-quality observations, rapid viewpoint changes, occlusions, and repetitive structures may overwrite reliable historical information, leading to memory contamination, catastrophic forgetting, and accumulated pose drift. More recently, test-time adaptation has emerged as a promising direction for improving long-sequence 3D reconstruction. TTT3R~\cite{chenttt3r} and TTSA3R~\cite{zheng2026ttsa3r}  leverage test-time training to adapt recurrent states online and forget persistent states according to temporal state evolution and spatial observation quality. These developments suggest that memory is not merely an auxiliary component for 3D foundation models, but a core mechanism for transforming them from short-horizon geometry predictors into long-term robotic world representations.

Nevertheless, directly using recurrent memory is still insufficient for scalable robotic mapping,localization, and rendering. First, a single global state is difficult to maintain over long trajectories, since local errors may propagate through the entire memory and corrupt subsequent reconstruction. Second, adaptive state update alone cannot resolve global inconsistency caused by long-term drift or loop revisiting. Third, most existing 3D foundation models mainly output geometric representations such as depth maps, pointmaps, or camera poses, while robotic applications increasingly require maps that are not only geometrically accurate but also visually renderable for inspection, simulation, teleoperation, and downstream embodied reasoning. Therefore, a practical 3D foundation world model should satisfy three requirements simultaneously: it should maintain adaptive long-term memory, enforce local-to-global consistency over large-scale environments, and decode the learned world state into both geometric and renderable representations.

In this paper, we present a memory-centric  3D foundation model for scalable localization, dense reconstruction, and Gaussian rendering. The core of our framework is an adaptive world memory mechanism that enables global 3D memory propagation across long image streams. Specifically, we introduce a dual-stage adaptive memory regulation strategy. The first stage employs gated recurrent spatial states to learn memory propagation and suppress irrelevant historical information. The second stage performs test-time temporal-spatial memory regulation, where temporal state evolution and spatial observation-state consistency jointly calibrate token-wise memory update, preservation, and forgetting. By coupling learned recurrent gating with test-time adaptive regulation, the proposed memory mechanism allows the model to preserve reliable long-term geometry while incorporating new observations when necessary.

To further scale the memory to large robotic environments, we organize the world state in a submap-aware manner. Each submap maintains a local adaptive memory, which bounds error accumulation and prevents the entire global state from being contaminated by local tracking failures. During online operation, a progressive mapping and tracking strategy refines camera poses from coarse foundation-model predictions and progressively updates the active local map. When revisited regions are detected, loop closure constraints are introduced and optimized through an SL(4)-based submap optimization backend, enforcing global consistency across submaps while preserving local reconstruction accuracy. This design connects neural memory propagation with classical SLAM consistency optimization, allowing the model to operate robustly over long trajectories and large-scale scenes.

Beyond dense geometric reconstruction, we further introduce a 3D Gaussian reconstruction head that decodes the shared adaptive memory into renderable Gaussian primitives. Unlike conventional 3DGS-SLAM methods that usually require accurate pose initialization, depth input, or per-scene optimization, our framework uses the 3D foundation memory as a generalizable prior to jointly support camera pose estimation, dense point-cloud reconstruction, and photorealistic Gaussian rendering from RGB streams. The resulting representation bridges geometric SLAM and neural rendering: pointmaps and poses provide metric structure for localization and mapping, while Gaussian primitives provide a high-fidelity renderable map for novel-view synthesis and visual scene understanding.

We validate the proposed framework on public benchmarks and self-collected real-world datasets acquired using multiple robotic platforms, including wheeled mobile robots, handheld devices, and quadruped robots. These datasets cover diverse motion patterns, viewpoints, and deployment conditions, providing a more comprehensive evaluation of long-term robotic world modeling than standard indoor benchmarks alone. Extensive experiments demonstrate that our method improves trajectory accuracy, reconstruction completeness, and rendering quality compared with existing 3D foundation reconstruction and SLAM baselines, while maintaining efficient online inference.

The main contributions of this work are summarized as follows:

\begin{itemize}
    \item \textbf{Adaptive world memory 3D foundation model.} We propose a stateful 3D foundation model that unifies camera localization, dense point-cloud reconstruction, and 3D Gaussian rendering through a shared adaptive world memory.

    \item \textbf{Dual-stage adaptive memory regulation.} We develop a memory update mechanism that couples gated recurrent spatial states with test-time temporal-spatial regulation, enabling token-wise memory propagation, preservation, and forgetting over long RGB streams.
    \item \textbf{Submap-aware scalable global consistency.} We introduce a submap-aware memory organization with progressive mapping and tracking, loop closure, and SL(4)-based optimization to achieve scalable local-to-global consistency in large environments.

    \item \textbf{Renderable Gaussian world representation.} We design a 3D Gaussian reconstruction head that decodes adaptive world memory into renderable Gaussian primitives, extending 3D foundation models from geometric prediction to photorealistic scene mapping.
    \item\textbf{Multi-platform real-world evaluation.} We collect and evaluate on real-world datasets captured by wheeled mobile robots, handheld platforms, and quadruped robots, demonstrating the robustness and practicality of the proposed model in diverse robotic scenarios.
\end{itemize}

\begin{figure*}[!tbp]
  \centering
  \includegraphics[width=\linewidth]{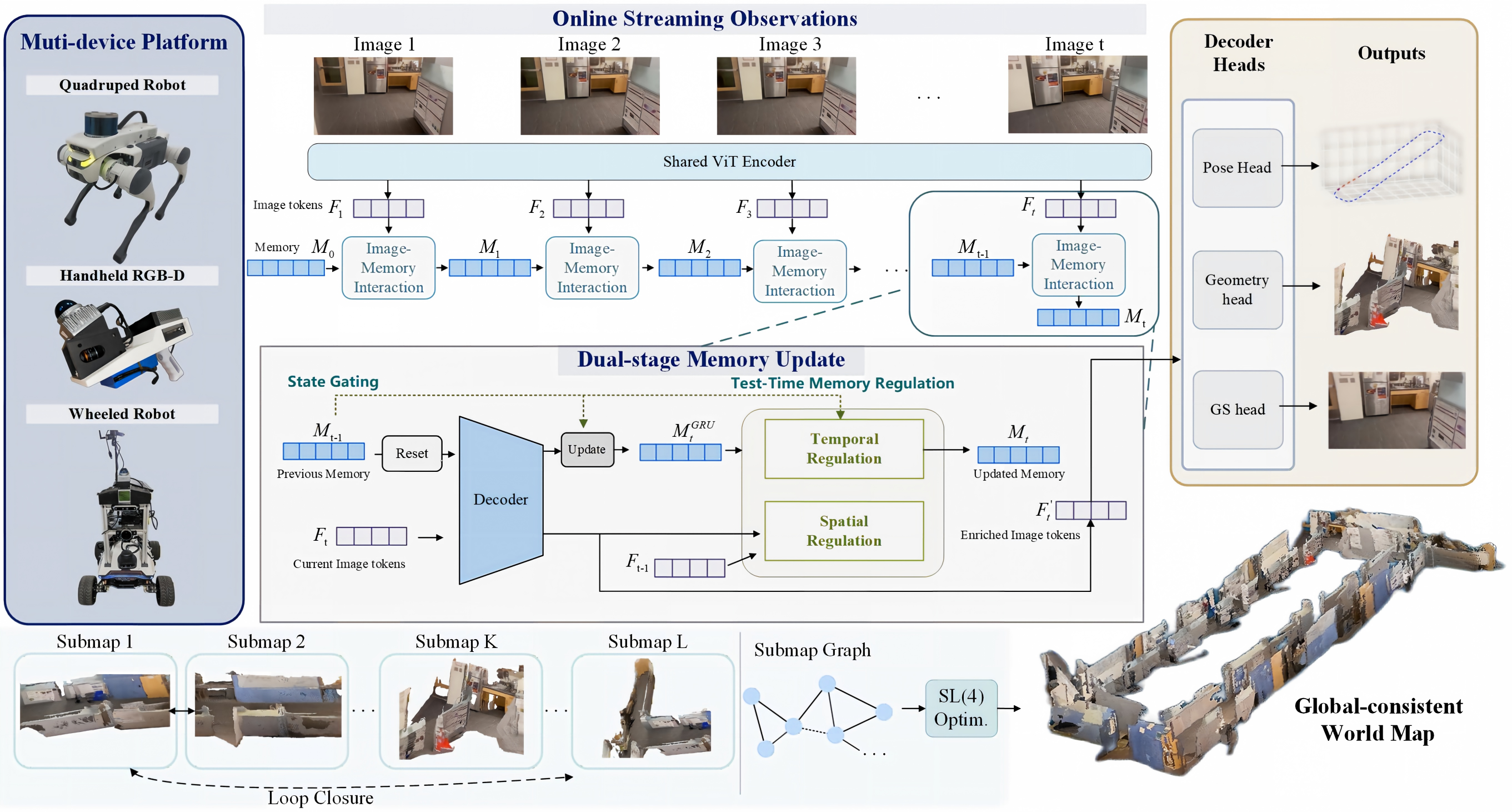}

  \caption{\textbf{Overall Framework}. For each incoming RGB frame, a shared 3D foundation encoder extracts geometry-aware visual tokens, which are propagated through an adaptive world memory regulated by a gated recurrent spatial state and test-time temporal-spatial regulation. The memory-enhanced features are decoded into camera poses, dense metric pointmaps with confidence maps, and renderable 3D Gaussian primitives. For large-scale robotic operation, the world memory is organized in a submap-aware manner, while loop closure and SL(4)-based submap optimization align local submaps into a globally consistent world frame.}
  \label{fig:framework}
\end{figure*}

\section{Related Works}

\subsection{Scene Representations for Mapping and Localization}
Scene representation is a central problem in robotic mapping and localization. 
Deng et al.~\cite{deng2025best3dscenerepresentation} provide a comprehensive survey of 3D scene representations for robotics, covering the evolution from classical geometric representations, such as sparse landmarks, dense point clouds, surfels, occupancy grids, and TSDF volumes, to neural scene representations, including NeRF~\cite{NeRF}, 3D Gaussian Splatting~\cite{3dgs}, and recent foundation-model-based representations. 
Traditional SLAM systems typically represent the environment using sparse hand-crafted visual landmarks. ORB-SLAM~\cite{orbslam2} and VINS~\cite{vins} relies on cisual keypoints for camera tracking and map construction,
Sparse landmark maps are computationally efficient and well suited for camera localization, but they provide limited geometric completeness and are difficult to use directly for dense scene understanding, obstacle avoidance, or high-fidelity visualization. 
Dense point clouds and surfel maps~\cite{kintinuous,elasticfusion} offer richer geometric structures and more interpretable scene maps, but they often suffer from redundancy, noise accumulation, incomplete surfaces, and limited visual realism.

Neural scene representations have provided new alternatives for dense mapping. NeRF-based SLAM methods represent scenes as continuous radiance fields and can produce high-quality novel-view synthesis~\cite{niceslam,eslam, plgslam,mneslam}. However, they usually require per-scene optimization and can be computationally expensive for online robotic deployment. More recently, 3D Gaussian Splatting~\cite{3dgs} has enabled explicit, efficient, and high-fidelity radiance-field rendering using Gaussian primitives. This has motivated a series of 3DGS-based SLAM systems~\cite{splatam,gsslam,photoslam,deng2024compact}, which bridge dense mapping and real-time rendering. Nevertheless, they typically rely on accurate camera poses, depth observations, or per-scene optimization, and therefore have limited generalization to unseen environments.

In parallel, 3D foundation models introduce a different type of scene representation. Instead of optimizing a map from scratch for each scene, they infer generalizable geometric representations such as pointmaps, depth maps, camera poses, and latent geometric tokens from RGB observations. This provides a promising direction for building robot maps with strong priors and reduced dependence on sensor-specific assumptions.

\subsection{3D Foundation models}

Recent 3D foundation models provide a new paradigm for generalizable geometric reasoning from RGB images. Instead of relying on hand-crafted correspondences, calibrated sensors, or scene-specific optimization, these models learn strong geometric priors from large-scale data and directly infer dense 3D structures from visual observations. DUSt3R~\cite{dust3r} introduced a pointmap-based formulation that regresses dense 3D geometry from image pairs without requiring known camera parameters. MASt3R~\cite{mast3r} further grounds image matching in 3D and improves correspondence estimation. More recent models, such as VGGT~\cite{vggt}, $\pi^3$~\cite{wang2025pi} and so on~\cite{wang2025amb3r}, extend this paradigm to multi-view inputs and jointly estimate camera poses, depth maps, and dense point clouds through global visual geometry reasoning. These models provide a powerful foundation for robotic 3D perception, since they reduce the dependence on carefully engineered matching pipelines and show strong generalization to unseen scenes. However, most existing 3D foundation models are still designed as feed-forward or short-horizon geometric predictors. As the sequence length grows, dense global attention and joint inference lead to increasing memory consumption and limited scalability. More importantly, these models usually lack a persistent world state that can be continuously updated during online robotic operation. In contrast, our work aims to transform a 3D foundation model from a short-horizon geometry predictor into a memory-centric world representation for long-term robotic perception.

\subsection{Online and Memory-based 3D Foundation Models}

To improve scalability, recent works have explored online and memory-based 3D foundation models. Instead of processing all frames simultaneously, streaming methods process observations sequentially and maintain compact historical information for online inference.
Efficiently handling long sequences and memory propagation have motivated the development of linearcomplexity architectures, such as Linear Transformers~\cite{katharopoulos2020transformers}, Mamba~\cite{gu2023mamba} DeltaNet~\cite{schlag2021linear}, and Test-Time Training (TTT)~\cite{sun2024learning}.
StreamVGGT~\cite{zhuo2025streaming} sparsifies global temporal reasoning through causal attention, reducing the cost of long-sequence encoding. Spann3R~\cite{spann3r} introduces an external spatial memory for incremental reconstruction, while CUT3R~\cite{cut3r} maintains recurrent state tokens to integrate sequential observations with a constant memory footprint. \cite{shen2025grs} shows that gated recurrent spatial states can improve dense SLAM by selectively propagating spatial memory.  Point3R~\cite{wu2026point3r} further explores explicit spatial pointer memory to aggregate local scene information. \cite{fang2026incvggt, yuan2026infinitevggt} use top-k most relevant/highest-scoring slots for history KV cache. Mamba-VGGT~\cite{deng2026mamba} introduces an external Mamba-based memory module to encode persistent global scene memory, thereby improving long-range geometric consistency and global coherence. These methods demonstrate that memory is essential for scaling 3D foundation models beyond short input sequences. Recent adaptive approaches attempt to use test-time adaption for online optimization. LaCT~\cite{Lact} dynamically updates non-linear MLP fast weights per token chunk. TTT3R~\cite{chenttt3r} and TTSA3R~\cite{zheng2026ttsa3r}  adopts test-time training to update recurrent states online based on temporal and spatial observation quality.  Further improvements in representation and scalability have been introduced by ZipMap\cite{jin2026zipmap}, LoGER~\cite{zhang2026loger}, Mem3R~\cite{liu2026mem3r}, Scal3r~\cite{xie2026scal3r}, and VGG-T$^3$~\cite{elflein2026vgg}. 
However, these methods mainly focus on geometric reconstruction and pose prediction, while the connection between adaptive foundation memory, renderable map representation, and SLAM-level global consistency remains under-explored. Our framework addresses this gap by coupling gated recurrent spatial states with test-time temporal-spatial regulation, and further organizing the memory into submaps for scalable localization, reconstruction, and rendering.

\subsection{Scalable SLAM Optimization and Global Consistency}
While learned 3D priors can provide strong local geometry and pose estimates, scalable robotic mapping still requires global consistency. Classical SLAM systems achieve this through tracking, keyframe selection, local bundle adjustment, loop closure, and pose-graph optimization~\cite{orbslam2,droidslam}. These components are essential for long-term operation, since local tracking errors inevitably accumulate over time and must be corrected when previously visited regions are revisited. Submap-based SLAM further improves scalability by dividing large environments into locally consistent maps and optimizing their relative transformations at the global level~\cite{plgslam}. This local-to-global design is particularly important for large-scale scenes, where directly optimizing all frames and map elements is computationally prohibitive. Recent learning-based SLAM and foundation-model-based reconstruction methods, such as MASt3R-SLAM~\cite{mast3rslam}, VGGT-SLAM~\cite{maggio2026vggt}, VGGT-Long~\cite{deng2025vggt} have begun to incorporate global alignment or optimization to improve consistency. However, many of them still focus on either local reconstruction or frame-wise pose prediction, and their learned memory is not explicitly coupled with loop closure and submap-level global correction. 

Our method follows the principle that adaptive neural memory and geometric SLAM optimization should be complementary. The proposed memory mechanism propagates and regulates global 3D information at the token level, while the submap-based backend enforces large-scale consistency at the map level.  This design connects memory-based 3D foundation modeling with scalable SLAM optimization, enabling robust localization, complete reconstruction, and renderable world modeling in long-sequence robotic environments.

\section{Method}
\subsection{Framework Overview}

Given a streaming monocular RGB sequence $\mathcal{I}=\{ I_t \}_{t=1}^{T}$, our goal is to build an online 3D world representation that simultaneously supports camera localization, dense reconstruction, and photorealistic rendering. The system takes RGB images as input without requiring depth measurements or a pre-built scene map. It maintains an adaptive world memory $\mathcal{M}$ and outputs camera poses $\mathcal{T}=\{ \mathbf{T}_t \}_{t=1}^{T}$, dense metric pointmaps $\mathcal{X}=\{ \mathbf{X}_t \}_{t=1}^{T}$ with confidence maps $\mathcal{C}=\{ \mathbf{C}_t \}_{t=1}^{T}$, and a renderable 3D Gaussian map $\mathcal{G}$.

The overview of our framework is shown in Fig.~\ref{fig:framework}. For each incoming RGB frame, a 3D foundation backbone first extracts geometry-aware visual tokens. These tokens are then propagated through an adaptive world memory (Sec.~\ref{sec:memory}), which is regulated by two complementary mechanisms: a gated recurrent spatial state for learned memory propagation and a test-time temporal--spatial regulation strategy for adaptive memory update, preservation, and forgetting. Based on the memory-enhanced features, the model predicts camera poses and dense metric pointmaps for localization and geometric reconstruction.

Beyond geometric outputs, we further introduce a Gaussian reconstruction head that decodes the visual tokens into renderable 3D Gaussian primitives (Sec.~\ref{sec:gaussian_world}). The predicted Gaussian primitives are incrementally fused into a submap-aware Gaussian world representation, enabling high-fidelity novel-view rendering together with online dense mapping. 

To support large-scale robotic operation, the world memory is further organized in a submap-aware manner (Sec.~\ref{sec:mapping}). Each submap maintains local memory and local scene representations, while progressive mapping and tracking jointly refine camera poses and update the active local map during online operation. When revisited regions are detected, loop closure and $\mathrm{SL}(4)$-based submap optimization are introduced to align local submaps into a globally consistent world frame. In this way, the proposed framework transforms a short-horizon 3D foundation model into a stateful world model for long-term localization, reconstruction, and rendering.

\subsection{Dual-Stage Adaptive World Memory}
\label{sec:memory}

Existing 3D foundation models usually process image pairs or short frame windows, which limits their ability to maintain long-term spatial consistency in online robotic scenarios. To address this issue, we introduce an adaptive world memory that progressively aggregates historical 3D information from streaming RGB observations. Different from uniform recurrent updates, our memory is regulated by a dual-stage mechanism. The first stage uses gated recurrent spatial states to learn how historical memory should be propagated and suppressed. The second stage performs test-time temporal-spatial regulation to further calibrate token-wise memory update, preservation, and forgetting according to online state evolution and observation quality. At each time step $t$, a shared 3D foundation encoder maps $I_t$ to geometry-aware visual tokens $\mathbf{F}_t \in \mathbb{R}^{K \times C}$, where $K$ and $C$ denote the token number and feature dimension. Inspired by CUT3R~\cite{cut3r}, we maintain a latent memory state $\mathbf{M}_{t-1} \in \mathbb{R}^{N \times C}$ that stores historical 3D information, where $N$ is the number of memory tokens. For simplicity, we omit the submap index in this subsection and describe the update within the active submap.

\subsubsection{Gated Recurrent Spatial State}

\begin{figure}[!t]
  \centering
  \includegraphics[width=\linewidth]{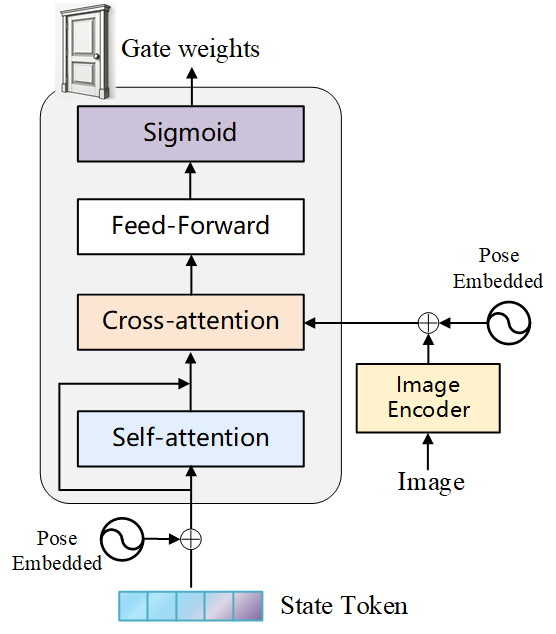}

  \caption{\textbf{Gate structure}. We illustrate the detailed structures of the reset and update gates we designed. Both gates take the historical latent state and the current frame as inputs, and compute the gating weights through a combination of self-attention and cross-attention mechanisms.}
  \label{fig:gate}
\end{figure}

Directly integrating new visual observations into the memory may introduce noisy or inconsistent information, especially under rapid viewpoint changes, occlusions, and low-overlap frames. Inspired by recurrent gated models~\cite{GRU,GRU2}, we use two transformer-based gates to regulate the interaction between the current visual tokens and the previous memory state. As shown in Fig.~\ref{fig:gate}, the image tokens $F_t$ will interact with the memory in two directions. We update the state with information from the current image feature, then we retrieve contextual cues from the state to incorporate knowledge accumulated from past frames. Specifically, a reset gate $G_r(\cdot)$ suppresses outdated or irrelevant historical information, while an update gate $G_u(\cdot)$ controls how much candidate memory should be integrated:
\begin{equation} \mathbf{R}_t = G_r(\mathbf{M}_{t-1}, \mathbf{F}_t), \quad \mathbf{U}_t = G_u(\mathbf{M}_{t-1}, \mathbf{F}_t), \end{equation} where $\mathbf{R}_t$ and $\mathbf{U}_t$ are token-wise gating weights. The reset gate is first applied to the previous memory: \begin{equation} \mathbf{M}^{\mathrm{reset}}_t = \mathbf{R}_t \odot \mathbf{M}_{t-1}. \end{equation} The reset memory, current visual tokens, and a pose token $\mathbf{z}_t$ are then fed into a transformer decoder: \begin{equation} [\hat{\mathbf{M}}_t,\mathbf{z}'_t \oplus \mathbf{F}'_t] = \mathrm{Decoder} \left( [\mathbf{M}^{\mathrm{reset}}_t,\mathbf{z}_t \oplus \mathbf{F}_t] \right), \end{equation} where $\hat{\mathbf{M}}_t$ denotes the candidate memory state, $\mathbf{F}'_t$ denotes memory-enhanced frame features, and $\mathbf{z}'_t$ is the updated pose token. Through cross-attention between memory tokens and image tokens, the decoder enables bidirectional information exchange: the current frame updates the memory, while the memory provides historical spatial context for the current prediction.

The memory can be updated as \begin{equation} \mathbf{M}^{\mathrm{gru}}_t = \mathbf{U}_t \odot \hat{\mathbf{M}}_t + (1-\mathbf{U}_t) \odot \mathbf{M}_{t-1}. \end{equation} However, this update is still determined by learned gates fixed after training, and may not sufficiently adapt to test-time variations in motion, viewpoint overlap, or observation quality. We therefore introduce a second-stage temporal--spatial regulation mechanism.

\subsubsection{Test-Time Temporal--Spatial Memory Regulation}

The goal of test-time regulation is to determine which memory tokens should be updated, preserved, or forgotten during online inference. Inspired by~\cite{zheng2026ttsa3r}, we compute an adaptive mask from two complementary cues: temporal state evolution and spatial observation-state consistency.

First, temporal regulation measures how candidate memory tokens evolve across consecutive timesteps. Tokens with small temporal changes usually correspond to stable and reliable geometry, while tokens with large changes indicate regions that require new observations or correction. We compute the token-wise state variation between two consecutive gated candidate states as \begin{equation} \Delta_t = \mathrm{Norm}(\mathbf{M}^{\mathrm{gru}}_t-\mathbf{M}^{\mathrm{gru}}_{t-1}) \in \mathbb{R}^{N}, \end{equation} where $\mathrm{Norm}(\cdot)$ denotes the L2 norm along the feature dimension. To make the variation comparable across different scenes and motions, we normalize it by the average token variation with a numerical stabilizer: \begin{equation} \bar{\Delta}_t = \frac{\Delta_t} {\frac{1}{N}\sum_{i=1}^{N}\Delta_t^{(i)}+\epsilon_{\Delta}} \in \mathbb{R}^{N}, \end{equation} where $\epsilon_{\Delta}>0$ prevents division by zero when all token variations vanish. The temporal adaptive mask is then obtained by sigmoid gating: \begin{equation} \mathbf{A}^{\mathrm{temp}}_t = \sigma\left(\bar{\Delta}_t-\tau_{\mathrm{temp}}\right) \in \mathbb{R}^{N}, \end{equation} where $\tau_{\mathrm{temp}}$ denotes the temporal-update sensitivity threshold. This mask encourages aggressive updates for rapidly changing tokens and preserves stable memory tokens for long-term consistency.

Second, spatial regulation evaluates whether the current observation provides useful spatial information for updating each memory token. We first compute the feature divergence between consecutive frames using cosine dissimilarity: \begin{equation} \mathbf{D}_t = 1-\mathrm{CosSim}(\mathbf{F}_t,\mathbf{F}_{t-1}) \in \mathbb{R}^{K}. \end{equation} Meanwhile, we aggregate the cross-attention maps between memory tokens and image tokens over the decoder layers and attention heads: \begin{equation} \mathbf{A}^{\mathrm{attn}}_t = \left|\frac{1}{LH}\sum_{l=1}^{L}\sum_{h=1}^{H} \mathrm{CrossAttn}^{(l,h)}_t[:,1:]\right| \in \mathbb{R}^{N\times K}, \end{equation} where $L$ and $H$ denote the numbers of decoder layers and attention heads, respectively. The cross-attention map before slicing has shape $N\times(K+1)$, and $[:,1:]$ removes the pose-token column. The spatial update mask is computed by combining attention alignment and feature divergence:
\begin{equation}
\begin{aligned}
\left[\mathbf{A}^{\mathrm{spat}}_t\right]^{(i)}
&= \sigma \left( \max_{1\leq k\leq K} \left( \left[\mathbf{A}^{\mathrm{attn}}_t\right]^{(i,k)} \mathbf{D}_t^{(k)} \right) \right),\\
\mathbf{A}^{\mathrm{spat}}_t &\in\mathbb{R}^{N}.
\end{aligned}
\end{equation}
Here, $\mathbf{D}_t$ is broadcast over the $N$ memory-token rows, and the maximum is taken along the image-token dimension $K$. This design activates memory updates only when a memory token strongly attends to image regions that contain meaningful visual changes, thereby avoiding unnecessary updates to already stable geometry. The final test-time adaptive mask is obtained by fusing temporal and spatial cues: \begin{equation} \mathbf{A}_t = \mathbf{A}^{\mathrm{temp}}_t \odot \mathbf{A}^{\mathrm{spat}}_t \in \mathbb{R}^{N}. \end{equation} During memory fusion, $\mathbf{A}_t$ is expanded to $N\times1$ and broadcast along the feature dimension $C$. For the first valid frame and after a sequence reset, temporal--spatial regulation is bypassed and the gated candidate state is directly accepted, which is equivalent to setting $\mathbf{A}_t=\mathbf{1}$.

\subsubsection{Dual-Stage Memory Update} We combine the learned recurrent update and the test-time adaptive mask in two consecutive stages. The learned update gate $\mathbf{U}_t$ is first used to obtain $\mathbf{M}^{\mathrm{gru}}_t$, as defined above. The adaptive world memory is then updated by \begin{equation} \mathbf{M}_t = \mathbf{A}_t \odot \mathbf{M}^{\mathrm{gru}}_t + (1-\mathbf{A}_t)\odot \mathbf{M}_{t-1}. \end{equation} The learned gate $\mathbf{U}_t$ provides data-driven recurrent memory propagation, while the adaptive mask $\mathbf{A}_t$ calibrates the resulting gated candidate state at test time according to temporal and spatial consistency. Therefore, the proposed dual-stage update prevents unreliable observations from overwriting long-term memory, while still allowing necessary geometric refinement when the scene or viewpoint changes. After the memory update, the memory-enhanced frame tokens $\mathbf{F}'_t$ and pose token $\mathbf{z}'_t$ are used for downstream prediction: \begin{equation} \hat{\mathbf{X}}^{\mathrm{self}}_t, \hat{\mathbf{C}}^{\mathrm{self}}_t = H_{\mathrm{self}}(\mathbf{F}'_t), \end{equation} \begin{equation} \hat{\mathbf{X}}^{\mathrm{world}}_t, \hat{\mathbf{C}}^{\mathrm{world}}_t = H_{\mathrm{world}}(\mathbf{F}'_t,\mathbf{z}'_t), \end{equation} \begin{equation} \hat{\mathbf{T}}_t = H_{\mathrm{pose}}(\mathbf{z}'_t). \end{equation} Here, $H_{\mathrm{self}}$ predicts pointmaps in the local camera coordinate, while $H_{\mathrm{world}}$ predicts pointmaps in the reference coordinate system of the current active submap; the superscript ``world'' distinguishes this output from the self-coordinate prediction and does not denote the final globally aligned frame. $H_{\mathrm{pose}}$ predicts the 6-DoF camera pose. The Gaussian reconstruction head is introduced later in Sec.~\ref{sec:gaussian_world}, where the same adaptive memory is further decoded into a renderable Gaussian world representation.

\subsection{Feed-forward Gaussian World Representation}
\label{sec:gaussian_world}

The adaptive world memory provides a persistent latent representation for long-term geometric reasoning. 
Beyond predicting metric pointmaps and camera poses, we further decode the same memory-enhanced features into a renderable Gaussian world representation. 
Different from conventional 3DGS-SLAM methods that initialize and optimize Gaussian primitives mainly through per-scene tracking and mapping, our Gaussian primitives are directly predicted from the 3D foundation memory and anchored by the metric pointmaps estimated by the geometric head.

As introduced in Sec.~\ref{sec:memory}, the memory-enhanced frame tokens $\mathbf{F}'_t$ and pose token $\mathbf{z}'_t$ are used to predict the pointmap $\hat{\mathbf{X}}^{\mathrm{world}}_t$, confidence map $\hat{\mathbf{C}}^{\mathrm{world}}_t$, and camera pose $\hat{\mathbf{T}}_t$. Within the submap-aware system, $\hat{\mathbf{X}}^{\mathrm{world}}_t$ is expressed in the reference coordinate system of the current active submap rather than the final globally aligned world frame.
We use this active-submap-coordinate pointmap as the geometric anchor of Gaussian primitives. For each valid pixel or sampled point $g$, the center of the corresponding Gaussian primitive is defined as
$
    \boldsymbol{\mu}_g = \hat{\mathbf{X}}^{\mathrm{world}}_t(g).
$
This design avoids additional depth back-projection and ensures that the Gaussian map is geometrically consistent with the metric pointmap representation.

To predict the remaining Gaussian attributes, we introduce a pixel-wise Gaussian parameter head. 
The head combines memory-enhanced geometric features with shallow appearance features extracted from the input RGB image. 
Specifically, a DPT-style decoder $F_d(\cdot)$ transforms $\mathbf{F}'_t$ into dense geometric features, while a lightweight CNN encoder $F_a(\cdot)$ extracts local appearance features from $I_t$. 
The fused feature is then fed into a regression head $F_b(\cdot)$:
\begin{equation}
    \{
    \sigma_g,
    \mathbf{r}_g,
    \mathbf{s}_g,
    \mathbf{c}_g
    \}_{g=1}^{G}
    =
    F_b
    \left(
    F_d(\mathbf{F}'_t)
    +
    F_a(I_t)
    \right),
\end{equation}
where $\sigma_g$ denotes opacity, $\mathbf{r}_g$ denotes rotation, $\mathbf{s}_g$ denotes anisotropic scale, and $\mathbf{c}_g$ denotes color or spherical harmonic coefficients. We associate each primitive with the detached geometric reliability of its active-submap-coordinate pointmap anchor:
\begin{equation}
    C^{\mathrm{geo}}_g
    =
    \operatorname{sg}\!\left(
    \hat{\mathbf{C}}^{\mathrm{world}}_t(g)
    \right),
\end{equation}
where $\operatorname{sg}(\cdot)$ denotes the stop-gradient operator. Thus, $C^{\mathrm{geo}}_g$ is metadata inherited from the geometry prediction rather than an independently learned Gaussian-head output, and no gradient from Gaussian rendering or map insertion is propagated through this score.
The covariance matrix of each Gaussian is computed from the predicted rotation and scale:
\begin{equation}
    \boldsymbol{\Sigma}_g
    =
    \mathbf{R}(\mathbf{r}_g)
    \mathrm{diag}(\mathbf{s}_g^2)
    \mathbf{R}(\mathbf{r}_g)^{\top}.
\end{equation}
Thus, each primitive predicted from $I_t$ is represented by $(\boldsymbol{\mu}_g,\boldsymbol{\Sigma}_g,\sigma_g,\mathbf{c}_g,C^{\mathrm{geo}}_g)$, and the collection of all $G$ primitives forms $\mathcal{G}_t$.

The detached geometric reliability scores are used to suppress primitives anchored at uncertain pointmap predictions before map insertion.
For the active submap $s_t$, we insert only Gaussian primitives whose geometric anchors are reliable:
\begin{equation}
    \mathcal{G}^{s_t}_{t}
    =
    \mathcal{G}^{s_t}_{t-1}
    \cup
    \{
    g \in \mathcal{G}_t
    \mid
    C^{\mathrm{geo}}_g > \tau_g
    \},
\end{equation}
where $\tau_g$ is the map-insertion threshold applied to the detached geometric confidence. This reliability-aware insertion strategy reduces noisy Gaussian primitives caused by uncertain geometry without introducing a separately optimized confidence variable.

For each input frame, the model outputs a camera intrinsic matrix in the pixel coordinate system of its inference image:
\begin{equation}
    \hat{\mathbf{K}}^{\mathrm{pred}}_t
    =
    \begin{bmatrix}
    \hat{f}_{x,t} & 0 & \hat{c}_{x,t}\\
    0 & \hat{f}_{y,t} & \hat{c}_{y,t}\\
    0 & 0 & 1
    \end{bmatrix}.
\end{equation}
Before Gaussian rasterization, the model output is transformed into the renderer's pixel coordinate system. Let $s_{x,t}$ and $s_{y,t}$ denote the resize factors from the model inference image to the resized image, and let $(o_{x,t},o_{y,t})$ be the top-left crop origin measured in the resized image. We use
\begin{equation}
    \mathbf{Q}^{\mathrm{rc}}_t
    =
    \begin{bmatrix}
    s_{x,t} & 0 & -o_{x,t}\\
    0 & s_{y,t} & -o_{y,t}\\
    0 & 0 & 1
    \end{bmatrix},
    \qquad
    \hat{\mathbf{K}}^{\mathrm{gs}}_t
    =
    \mathbf{Q}^{\mathrm{rc}}_t\hat{\mathbf{K}}^{\mathrm{pred}}_t.
\end{equation}
Equivalently, the transformed intrinsic components are
\begin{equation}
\begin{aligned}
    f^{\mathrm{gs}}_{x,t} &= s_{x,t}\hat{f}_{x,t},
    & f^{\mathrm{gs}}_{y,t} &= s_{y,t}\hat{f}_{y,t},\\
    c^{\mathrm{gs}}_{x,t} &= s_{x,t}\hat{c}_{x,t}-o_{x,t},
    & c^{\mathrm{gs}}_{y,t} &= s_{y,t}\hat{c}_{y,t}-o_{y,t}.
\end{aligned}
\end{equation}
If Gaussian rasterization uses the model's inference resolution without an additional crop, $\mathbf{Q}^{\mathrm{rc}}_t=\mathbf{I}_3$, avoiding a duplicate intrinsic transformation.

Given the predicted camera pose $\hat{\mathbf{T}}_t$ and resize/crop-adjusted model-predicted intrinsics $\hat{\mathbf{K}}^{\mathrm{gs}}_t$, the local Gaussian map is rendered through differentiable Gaussian splatting:
\begin{equation}
    \hat{\mathbf{I}}_t
    =
    \mathcal{R}
    \left(
    \mathcal{G}^{s_t}_{t},
    \hat{\mathbf{T}}_t,
    \hat{\mathbf{K}}^{\mathrm{gs}}_t
    \right).
\end{equation}
where $\mathcal{R}(\cdot)$ denotes the Gaussian rasterization function. The transformed model output $\hat{\mathbf{K}}^{\mathrm{gs}}_t$ is used consistently throughout Gaussian rendering.
The rendered image provides a photometric signal for optimizing Gaussian attributes.

In this way, the proposed Gaussian reconstruction head converts the adaptive foundation memory into a renderable world representation. 
The pointmap head provides metric geometric anchors, the pose head places observations in the active-submap coordinate system, and the Gaussian reconstruction head predicts appearance-aware renderable primitives. 
Therefore, localization, dense reconstruction, and photorealistic rendering are unified through the same memory-enhanced 3D foundation representation.

\subsection{Progressive Mapping and Tracking}
\label{sec:mapping}

The proposed progressive mapping and tracking module links frame-wise foundation-model predictions to persistent local mapping and global submap alignment. For each incoming frame, the model predicts a camera pose, world-coordinate pointmap, confidence map, and Gaussian primitives. Starting from the predicted pose, the current pointmap is aligned with the active local map for geometric consistency. The refined pose is then used to insert reliable pointmap and Gaussian predictions into the active submap, allowing camera tracking and map construction to progress jointly online.

\subsubsection{Submap-Aware Local Mapping and Tracking}

Although adaptive world memory propagates historical 3D information, a single global memory over a long sequence remains vulnerable to drift and contamination. Long-range motion, abrupt viewpoint changes, repeated structures, and loop revisits may cause tracking failures or low-overlap observations to corrupt the entire state. We therefore organize the world memory into local submaps for scalability and robustness:
\begin{equation} \mathcal{S} = \{\mathcal{S}^{1}, \mathcal{S}^{2}, \ldots, \mathcal{S}^{S}\}, \end{equation} where $S$ denotes the number of constructed submaps. Each submap maintains its own adaptive memory state, local keyframes, camera poses, pointmaps, and Gaussian primitives: \begin{equation} \mathcal{S}^{s} = \{ \mathbf{M}^{s}, \mathcal{K}^{s}, \mathcal{T}^{s}, \mathcal{X}^{s}, \mathcal{C}^{s}, \mathcal{G}^{s}, \mathbf{T}^{w}_{s} \}, \end{equation}

where $\mathbf{M}^{s}$ is the local adaptive memory, $\mathcal{K}^{s}$ contains its keyframes, $\mathcal{T}^{s}$ and $\mathcal{X}^{s}$ are the local camera poses and pointmaps, $\mathcal{C}^{s}$ is the confidence map set, $\mathcal{G}^{s}$ contains the local Gaussian primitives, and $\mathbf{T}^{w}_{s}$ maps the submap coordinate to the global world frame. Each submap thus stores a geometrically coherent local memory and scene representation.

At time $t$, the system maintains an active submap $s_t$. Frame $I_t$ interacts with $\mathbf{M}^{s_t}_{t-1}$, and the dual-stage update in Sec.~\ref{sec:memory} is confined to this submap. The predicted pose initializes local tracking and is refined by confidence-weighted pointmap alignment with the active submap. After tracking, the memory, pointmap, camera pose, and reliable Gaussian primitives are incorporated into $\mathcal{S}^{s_t}$. Inactive submaps remain unchanged and persistent for loop closure and global optimization, preventing unreliable local updates from propagating through the world memory.

We use a keyframe policy to decide whether the current frame updates the active submap or initializes a new one. Let $I_k$ be the last keyframe and $I_{a_s}$ the anchor frame of the active submap; their overlap with the current frame is measured by $\mathrm{Cov}(\cdot,\cdot)$. Frame $I_t$ is inserted as a keyframe when $\mathrm{Cov}(I_t,I_k)<\tau_{\mathrm{kf}}$, where $\tau_{\mathrm{kf}}$ controls the insertion frequency. A new submap is initialized when $\mathrm{Cov}(I_t,I_{a_s})<\tau_{\mathrm{anchor}}$. The new submap uses $I_t$ as its anchor and initializes its memory from the observation and a learnable empty prior: \begin{equation} \mathbf{M}^{S+1}_{t} = \mathrm{InitMemory}(I_t, \mathbf{M}_0). \end{equation} Its transformation $\mathbf{T}^{w}_{S+1}$ is initialized from the tracking result relative to the previous active submap, connecting the adjacent submaps.

This submap-aware memory management provides three advantages. First, it bounds recurrent memory propagation within local regions, reducing long-range state degradation. Second, it prevents low-overlap or unreliable observations from contaminating previously established memories. Third, it provides a structured graph of local memories, which serves as the basis for loop closure and global submap optimization. Therefore, the proposed memory organization bridges token-level adaptive memory update and map-level global consistency.

\subsubsection{Loop Closure and SL(4)-Based Global Optimization}
\label{sec:tracking}

Independent local memories and maps bound local drift, but long-term operation still requires global correction for accumulated drift and revisited regions. We therefore construct a submap graph and perform loop-aware optimization over its transformations. Rather than restricting alignment to $\mathrm{SE}(3)$ or $\mathrm{Sim}(3)$, we optimize on the $\mathrm{SL}(4)$ manifold, providing flexibility to correct residual scale, shear, stretch, and projective distortions across independently reconstructed submaps.

Each submap $\mathcal{S}^{s}$ has a homogeneous transformation $\mathbf{H}_{s} \in \mathrm{SL}(4)$ from its local coordinates to the global world frame:
$
    \tilde{\mathbf{x}}^{w}
    =
    \mathbf{H}_{s}
    \tilde{\mathbf{x}}^{s},
$
where $\tilde{\mathbf{x}}$ denotes a homogeneous 3D point. The submap graph and its edge set are
\begin{equation} \mathcal{G}_{\mathrm{sub}}=(\mathcal{V},\mathcal{E}), \quad \mathcal{E}=\mathcal{E}_{\mathrm{seq}}\cup\mathcal{E}_{\mathrm{loop}}. \end{equation}
Each node $v_s\in\mathcal{V}$ represents a submap $\mathcal{S}^{s}$, and each edge $(i,j)\in\mathcal{E}$ carries a relative constraint $\bar{\mathbf{H}}_{ij}\in\mathrm{SL}(4)$. The graph contains sequential constraints between adjacent submaps and loop constraints between non-adjacent overlapping submaps.

\paragraph{Relative Submap Alignment.}

For overlapping submaps $\mathcal{S}^{i}$ and $\mathcal{S}^{j}$, we estimate a relative homography $\bar{\mathbf{H}}_{ij}$ from corresponding local points $\mathbf{x}^{i}_{a}$ and $\mathbf{x}^{j}_{b}$. Their ideal alignment satisfies
$
    \tilde{\mathbf{x}}^{i}_{a}
    =
    \bar{\mathbf{H}}_{ij}
    \tilde{\mathbf{x}}^{j}_{b}.
$
For adjacent submaps, dense correspondences come from overlapping keyframes shared during construction. For loop closures, retrieved historical frames are inserted into the current local inference window so that the same observations are reconstructed in both coordinate systems. This yields dense cross-submap correspondences without sparse feature matching.

Given the homogeneous point pairs, the relative homography is estimated by solving
$
    \mathbf{A}\mathbf{h}=0,
$
where $\mathbf{h}\in\mathbb{R}^{16}$ vectorizes the $4\times4$ homography. We use a RANSAC minimal solver for robustness to noisy pointmaps and outliers, then normalize the estimate to unit determinant:
\begin{equation}
    \bar{\mathbf{H}}_{ij}
    \leftarrow
    \frac{
    \bar{\mathbf{H}}_{ij}
    }{
    \det(\bar{\mathbf{H}}_{ij})^{1/4}
    },
    \quad
    \bar{\mathbf{H}}_{ij}\in \mathrm{SL}(4).
\end{equation}
This normalization places the relative constraint on the $\mathrm{SL}(4)$ manifold.

\paragraph{Loop Closure Detection.}

We use UniPR-3D~\cite{unipr-3d} descriptors for keyframes in previous submaps and retrieve candidate loop frames from non-adjacent historical submaps before finalizing a new submap. Given $I_a\in\mathcal{K}^{s}$ and $I_b\in\mathcal{K}^{i}$, a loop candidate is accepted only when descriptor retrieval and geometric verification satisfy
\begin{equation} \mathrm{sim}(d(I_a),d(I_b))>\tau_{\mathrm{loop}}, \quad N_{\mathrm{inlier}}>\tau_{\mathrm{inlier}}, \quad e_{\mathrm{geo}}<\tau_{\mathrm{geo}}. \end{equation}
Here, $d(\cdot)$ denotes the image descriptor. Retrieved loop frames provide cross-submap correspondences for estimating $\bar{\mathbf{H}}_{is}$, while the inlier and residual tests reject false closures caused by perceptual aliasing or repeated structures.

\paragraph{Global Optimization on the SL(4) Manifold.}

Given the sequential and loop constraints, we estimate the absolute submap transformations $\{\mathbf{H}_{s}\}_{s=1}^{S}$ with a nonlinear factor-graph objective on $\mathrm{SL}(4)$:
\begin{equation}
    \{\mathbf{H}^{*}_{s}\}_{s=1}^{S}
    =
    \arg\min_{\mathbf{H}_{s}\in\mathrm{SL}(4)}
    \sum_{(i,j)\in\mathcal{E}}
    \left\|
    \mathrm{Log}
    \left(
    \mathbf{H}^{-1}_{i}
    \mathbf{H}_{j}
    \bar{\mathbf{H}}^{-1}_{ij}
    \right)
    \right\|^{2}_{\boldsymbol{\Omega}_{ij}},
\end{equation}
where $\mathrm{Log}(\cdot)$ maps an $\mathrm{SL}(4)$ element to its 15-dimensional Lie algebra and $\boldsymbol{\Omega}_{ij}$ is the constraint information matrix. We fix the gauge by anchoring the first submap and iteratively applying tangent-space updates:
\begin{equation} \mathbf{H}_{1}=\mathbf{I}, \quad \mathbf{H}_{s}\leftarrow\mathbf{H}_{s}\mathrm{Exp}(\boldsymbol{\xi}_{s}), \quad \boldsymbol{\xi}_{s}\in\mathbb{R}^{15}. \end{equation}

The optimized transformations are applied to camera poses, pointmaps, and Gaussian primitives. A point or Gaussian center $\boldsymbol{\mu}^{s}$ is globally aligned by
$
    \tilde{\boldsymbol{\mu}}^{w}
    =
    \mathbf{H}^{*}_{s}
    \tilde{\boldsymbol{\mu}}^{s}.
$
Gaussian covariance is transformed using the local affine approximation around its center:
$
    \boldsymbol{\Sigma}^{w}
    =
    \mathbf{J}_{s}
    \boldsymbol{\Sigma}^{s}
    \mathbf{J}_{s}^{\top},
$
where $\mathbf{J}_{s}$ is the Jacobian of the inhomogeneous projection induced by $\mathbf{H}^{*}_{s}$. Thus, loop closure and global optimization jointly correct the geometric map, camera trajectory, and Gaussian world representation.

The backend complements the adaptive memory front-end: dual-stage updates maintain local temporal consistency at the token level, while $\mathrm{SL}(4)$ optimization enforces global consistency at the map level. Together, they enable scalable long-sequence localization, dense reconstruction, and renderable world modeling.

\subsection{Training Objectives}
\label{sec:training}

We train the proposed framework with geometric, pose, rendering, and regularization objectives. 
The geometric and pose losses supervise the 3D foundation backbone and memory-based prediction heads, while the Gaussian reconstruction head is trained without requiring ground-truth Gaussian primitives. 
Instead, the predicted Gaussian primitives are supervised through differentiable rendering using RGB reconstruction losses. 
In addition, during online inference, we use a progressive mapping and tracking objective to refine camera poses and update the active local map.

\paragraph{Pointmap regression loss.}
Following prior pointmap-based 3D foundation models~\cite{shen2025grs,cut3r}, we supervise dense metric pointmap prediction with a confidence-weighted 3D regression loss. 
Let $\mathbf{x}_i$ be the ground-truth 3D point at pixel $i$, $\hat{\mathbf{x}}_i$ be the predicted point, and $c_i\in[0,1]$ be the predicted confidence. 
We define the scale-aware regression loss as
\begin{equation}
    \mathcal{L}_{\mathrm{regr}}
    =
    \sum_{i=1}^{M}
    \left(
    c_i
    \left\|
    \frac{\hat{\mathbf{x}}_i}{\hat{s}}
    -
    \frac{\mathbf{x}_i}{s}
    \right\|_2
    -
    \beta \log c_i
    \right),
\end{equation}
where $s$ and $\hat{s}$ denote the normalization factors of the ground-truth and predicted pointmaps, respectively. 
When the training data provide metric-scale pointmaps, we set $\hat{s}=s$ to remove scale ambiguity. 
This loss is applied to both self-coordinate and world-coordinate pointmaps:
\begin{equation}
    \mathcal{L}_{\mathrm{geo}}
    =
    \mathcal{L}_{\mathrm{regr}}
    \left(
    \hat{\mathbf{X}}^{\mathrm{self}},
    \mathbf{X}^{\mathrm{self}}
    \right)
    +
    \lambda_{\mathrm{world}}
    \mathcal{L}_{\mathrm{regr}}
    \left(
    \hat{\mathbf{X}}^{\mathrm{world}},
    \mathbf{X}^{\mathrm{world}}
    \right).
\end{equation}

\paragraph{Pose loss.}
Let the predicted camera pose at time $t$ be $\hat{\mathbf{P}}_t=(\hat{\mathbf{q}}_t,\hat{\boldsymbol{\tau}}_t)$, where $\hat{\mathbf{q}}_t$ and $\hat{\boldsymbol{\tau}}_t$ denote the predicted quaternion and translation. 
Given the ground-truth pose $\mathbf{P}_t=(\mathbf{q}_t,\boldsymbol{\tau}_t)$, we minimize the rotational and translational discrepancies:
\begin{equation}
    \mathcal{L}_{\mathrm{pose}}
    =
    \sum_{t=1}^{N}
    \left(
    \left\|
    \hat{\mathbf{q}}_t-\mathbf{q}_t
    \right\|_2
    +
    \lambda_{\tau}
    \left\|
    \frac{\hat{\boldsymbol{\tau}}_t}{\hat{s}}
    -
    \frac{\boldsymbol{\tau}_t}{s}
    \right\|_2
    \right).
\end{equation}
The same normalization factor as in the pointmap regression loss is used for translation supervision.

\paragraph{Gaussian rendering loss.}
The Gaussian reconstruction head does not require ground-truth Gaussian parameters. Instead, the predicted primitives are supervised by rendering $\mathcal{G}_t$ with the predicted pose $\hat{\mathbf{T}}_t$ and resize/crop-adjusted model-predicted intrinsics $\hat{\mathbf{K}}^{\mathrm{gs}}_t$, i.e., $\hat{\mathbf{I}}_t=\mathcal{R}(\mathcal{G}_t,\hat{\mathbf{T}}_t,\hat{\mathbf{K}}^{\mathrm{gs}}_t)$. The rendering loss combines photometric, structural, and perceptual terms:
\begin{equation}
    \mathcal{L}_{\mathrm{rgb}}
    =
    \left\|
    \hat{\mathbf{I}}_t-\mathbf{I}_t
    \right\|_1
    +
    \lambda_{\mathrm{ssim}}
    \frac{1-\mathrm{SSIM}(\hat{\mathbf{I}}_t,\mathbf{I}_t)}{2}
    +
    \lambda_{\mathrm{perc}}
    \mathcal{L}_{\mathrm{perc}}
    (\hat{\mathbf{I}}_t,\mathbf{I}_t).
\end{equation}
Here, $\mathcal{L}_{\mathrm{perc}}$ denotes a perceptual loss. 
This rendering supervision allows the Gaussian reconstruction head to learn opacity, scale, rotation, and color attributes without explicit 3D Gaussian annotations.

\paragraph{Gaussian regularization.}
To stabilize Gaussian prediction and suppress degenerate primitives, we further apply regularization on opacity and scale:
\begin{equation}
    \mathcal{L}_{\mathrm{gs}}
    =
    \lambda_{\sigma}
    \mathcal{L}_{\sigma}
    +
    \lambda_{s}
    \mathcal{L}_{s}.
\end{equation}
The scale regularization penalizes overly large or collapsed Gaussians:
\begin{equation}
    \mathcal{L}_{s}
    =
    \sum_{g}
    \left(
    \max(0, \|\mathbf{s}_g\|_2-s_{\max})
    +
    \max(0, s_{\min}-\|\mathbf{s}_g\|_2)
    \right),
\end{equation}
where $s_{\min}$ and $s_{\max}$ define the valid scale range, and the opacity term is $\mathcal{L}_{\sigma}=\sum_g|\sigma_g|$. The detached reliability score $C^{\mathrm{geo}}_g$ is used only for map insertion and is therefore not included in $\mathcal{L}_{\mathrm{gs}}$.

\paragraph{Overall training objective.}
The full training objective is defined as
\begin{equation}
    \mathcal{L}_{\mathrm{train}}
    =
    \mathcal{L}_{\mathrm{geo}}
    +
    \lambda_{\mathrm{pose}}
    \mathcal{L}_{\mathrm{pose}}
    +
    \lambda_{\mathrm{rgb}}
    \mathcal{L}_{\mathrm{rgb}}
    +
    \lambda_{\mathrm{gs}}
    \mathcal{L}_{\mathrm{gs}}.
\end{equation}
This objective jointly trains the adaptive memory, geometric prediction heads, pose head, and Gaussian reconstruction head.

\paragraph{Curriculum training.}
We adopt a curriculum training strategy for stable convergence. 
First, we train on short sequences and freeze the image encoder and most decoder layers, updating only the memory regulation modules and task heads. 
Second, we unfreeze the decoder and train it jointly with the gated memory modules and prediction heads. 
Finally, we extend the training sequence length and fine-tune the full model, including the adaptive memory, pose head, pointmap heads, and Gaussian reconstruction head. 
This curriculum gradually strengthens inter-frame reasoning, long-range memory propagation, and renderable scene modeling.

\section{Experiments}

We evaluate the proposed method from four aspects: camera pose estimation accuracy, depth estimation, dense 3D reconstruction quality, Gaussian rendering quality. To comprehensively assess the scalability and robustness of the proposed memory-centric 3D foundation model, we conduct experiments on both public benchmarks and self-collected real-world datasets captured by different robotic platforms.

\subsection{Experiments Setup}
\subsubsection{Test Datasets.} We evaluate the proposed method on a broad set of public benchmarks covering depth estimation, camera pose estimation, dense reconstruction, Gaussian rendering, dynamic scenes, and large-scale long-sequence scenarios. For video depth estimation, we use Sintel~\cite{sintel}, Bonn~\cite{bonn}, KITTI~\cite{kitti}, and Replica~\cite{replica} under per-sequence and metric-scale protocols. Camera pose estimation is evaluated on 7-Scenes~\cite{7scenes}, TUM RGB-D~\cite{tum}, Sintel, TUM-dynamics~\cite{tum}, ScanNet~\cite{scannet}, KITTI Odometry~\cite{Geiger2012CVPR}, and Replica, covering static indoor scenes, dynamic content, and long trajectories. For dense reconstruction, we use 7-Scenes, NRGBD~\cite{neural_rgbd}, Replica/Apartment, and the self-collected indoor, outdoor, and indoor--outdoor sequences. Gaussian rendering is evaluated on ScanNet and BundleFusion~\cite{Bundle-fusion}, together with large-scale and real-world sequences. We further use Apartment and the self-collected dataset to assess long-range memory propagation, submap optimization, loop closure, and large-scale reconstruction and rendering consistency.

\subsubsection{Self-Collected Multi-Platform Dataset.}In addition to public benchmarks, we collect a real-world multi-platform dataset to evaluate the practicality of our method under diverse robotic deployment conditions. The dataset contains RGB sequences captured by three types of platforms: handheld devices, wheeled mobile robots, and quadruped robots, as summarized in Table~\ref{tab:dataset_summary}. The collected scenes include office rooms, long corridors, multi-room indoor environments, and indoor--outdoor hybrid scenarios. To further increase scene diversity, the data are collected in different geographic locations, including Singapore and Shanghai. Compared with standard indoor benchmarks, this dataset contains more diverse motion patterns, viewpoint changes, platform-induced motion disturbances, illumination variations, and scene layouts. It therefore provides a challenging testbed for evaluating whether the proposed adaptive world memory can generalize across different physical platforms and real robotic operating conditions. For quantitative trajectory evaluation, the reference trajectories are obtained using a LiDAR-based FAST-LIO system and transformed to the camera coordinate system through temporally synchronized and spatially calibrated LiDAR--camera measurements. The LiDAR measurements are used only to generate the reference trajectories for evaluation, while the proposed method takes only RGB images as input.

\begin{table*}[!tbp]
	\centering
	\scalebox{0.85}{
		\setlength{\tabcolsep}{2mm}
		\begin{tabular}{llcccc}
			\toprule
			Scene & Platform & Loop & Number of Frames & Indoor/Outdoor & Description \\
			\midrule
			\rowcolor{C7!50}Front Door & Quadruped Robot & \color{ForestGreen}{\cmark} & 4116 & Outdoor & Outdoor mixed terrain with steps, lawn, and asphalt pavement \\
			\rowcolor{C7!50}Garden & Quadruped Robot & \color{ForestGreen}{\cmark} & 2615 & Outdoor & Outdoor rose garden with stone paths, steps, and dense floral vegetation \\
			\rowcolor{C7!50}2F Circle with pp & Wheeled Robot & \color{ForestGreen}{\cmark} & 2214 & Indoor & Indoor circular corridor with moving pedestrians (dynamic interference) \\
			\rowcolor{C7!50}2F Circle without pp & Wheeled Robot & \color{ForestGreen}{\cmark} & 2139 & Indoor & Static indoor circular corridor with no pedestrians \\
			\rowcolor{C7!50}2F Long Loop & Wheeled Robot & \color{ForestGreen}{\cmark} & 2007 & Indoor & Indoor long rectangular corridor loop with extended path \\
			\rowcolor{C7!50}3F Office & Wheeled Robot & \color{ForestGreen}{\cmark} & 1986 & Indoor & Open-plan indoor office space with workstations \\
			\rowcolor{C7!50}1F Circle & Wheeled Robot & \color{ForestGreen}{\cmark} & 1685 & Indoor & Indoor circular lobby with an open central area \\
			\rowcolor{C7!50}Dark Night Road 1 & Wheeled Robot & \color{BrickRed}{\xmark} & 10081 & Outdoor &  Outdoor road in extremely low-light night conditions (high challenge)\\
			\rowcolor{C7!50}Dark Night Road 2 & Wheeled Robot & \color{ForestGreen}{\cmark} & 7012 & Outdoor & Outdoor road in dim night with faint ambient illumination \\
			\rowcolor{C7!50}Mid Fir Floor Mono & Handheld RGB-D& \color{ForestGreen}{\cmark} & 3889 & Both & Challenging mixed indoor-outdoor scene with varying structures \\
			\rowcolor{C7!50}Mid Indoor Mono & Handheld RGB-D& \color{ForestGreen}{\cmark} & 2157 & Indoor & Typical indoor office workspace scene \\
			\rowcolor{C7!50}Mid Indoor Mono 2 & Handheld RGB-D& \color{ForestGreen}{\cmark} & 1302 & Indoor & Compact indoor office environment with furniture layout \\
			\rowcolor{C7!50}Mid Sec Floor Mono & Handheld RGB-D& \color{ForestGreen}{\cmark} & 1882 & Indoor & Indoor circular corridor on the second floor \\
			\rowcolor{C7!50}216 & Quadruped Robot & \color{ForestGreen}{\cmark} & 783 & Indoor & Small enclosed indoor office room \\
			\rowcolor{C7!50}e\_valley & Quadruped Robot & \color{ForestGreen}{\cmark} & 923 & Indoor & 	Narrow indoor linear corridor \\
			\rowcolor{C7!50}passage4\_5 & Quadruped Robot & \color{ForestGreen}{\cmark} & 923 & Indoor & Challenging long straight indoor corridor with homogeneous texture\\
			\bottomrule
		\end{tabular}
	}
	\caption{Summary of dataset characteristics including loop presence, frame count, scene type, and platform.}
	\label{tab:dataset_summary}
\end{table*}

\subsubsection{Evaluation Metrics.} For video depth estimation, we report the absolute relative error (Abs Rel) and the percentage of pixels with relative error below $1.25$, denoted as $\delta < 1.25$. For camera pose estimation, we report the root mean square error of absolute trajectory error, denoted as RMSE of ATE, relative translation error ($RPE_{\mathrm{trans}}$), and relative rotation error ($RPE_{\mathrm{rot}}$). Both RPE metrics are computed as RMSE over all adjacent evaluated frame pairs ($\Delta=1$ frame). For dense reconstruction, we report accuracy and completeness. Accuracy is computed as the mean nearest-neighbor distance from the reconstructed point cloud to the ground-truth point cloud, whereas completeness is computed in the opposite direction. Following NICER-SLAM~\cite{nicerslam}, Spann3R~\cite{spann3r}, and SLAM3R~\cite{slam3r}, both metrics are calculated without distance truncation and reported in centimeters. We generate ground-truth point clouds by projecting depth maps into 3D using known camera intrinsics and ground-truth poses for each test sequence. To account for potential scale discrepancies across different methods, we adopt the alignment strategy used in SLAM3R~\cite{slam3r}: a global similarity transformation is first estimated using the Umeyama algorithm, followed by ICP refinement to minimize residual geometric error. For Gaussian rendering evaluation, we report standard novel-view synthesis metrics, including PSNR, SSIM, and LPIPS, on held-out views when available. For online efficiency, we consider end-to-end throughput, module-wise latency, and peak GPU memory consumption.

\subsubsection{Implementation Details.} All training experiments are conducted on eight NVIDIA A100 GPUs with 80 GB memory each. Inference and online SLAM experiments are conducted on a single NVIDIA RTX 4090 GPU. Unless otherwise specified, all input images are resized to a fixed resolution for fair comparison across methods. The resize factors and crop offsets are retained to construct $\mathbf{Q}^{\mathrm{rc}}_t$ before Gaussian rasterization. For online evaluation, the model processes frames sequentially and maintains submap-aware adaptive memory without accessing future frames. The SL(4)-based submap optimization is triggered when a new submap is created or when loop closure constraints are detected.

\subsection{Video Depth Estimation}

We first evaluate the dense geometric prediction capability of our method on video depth estimation. Table~\ref{tab:depth} reports the results on Sintel, Bonn, and KITTI under both per-sequence scale alignment and metric-scale evaluation. We compare with optimization-based methods, full-attention 3D foundation models, and streaming reconstruction methods.

\begin{table*}[!tbp]
\centering
\setlength{\tabcolsep}{3pt}
\begin{tabular}{ccc|cc|cc|cc}
\toprule
\multirow{2}{*}{Alignment} & \multirow{2}{*}{Method} & \multirow{2}{*}{Type}
& \multicolumn{2}{c|}{Sintel} & \multicolumn{2}{c|}{Bonn} & \multicolumn{2}{c}{KITTI} \\
& & & Abs Rel $\downarrow$ & $\delta<1.25$ $\uparrow$
& Abs Rel $\downarrow$ & $\delta<1.25$ $\uparrow$
& Abs Rel $\downarrow$ & $\delta<1.25$ $\uparrow$ \\
\midrule
\multirow{12}{*}{Per-sequence scale}
& DUST3R-GA \cite{dust3r} & Optim & 0.656 & 45.2 & 0.155 & 83.3 & \cellcolor{tabthird}0.144 & \cellcolor{tabthird}81.3 \\
& MASt3R-GA \cite{mast3r} & Optim & 0.641 & 43.9 & 0.252 & 70.1 & 0.183 & 74.5 \\
& MonST3R-GA \cite{zhang2025monst3r} & Optim & \cellcolor{tabthird}0.378 & \cellcolor{tabthird}55.8 & \cellcolor{tabthird}0.067 & \cellcolor{tabthird}96.3 & 0.168 & 74.4 \\
& Easi3R \cite{chen2025easi3r} & FA & \cellcolor{tabsecond}0.377 & \cellcolor{tabsecond}55.9 & \cellcolor{tabsecond}0.059 & \cellcolor{tabsecond}97.0 & \cellcolor{tabsecond}0.102 & \cellcolor{tabsecond}91.2 \\
& VGGT \cite{vggt} & FA & \cellcolor{tabfirst}0.287 & \cellcolor{tabfirst}66.1 & \cellcolor{tabfirst}0.055 & \cellcolor{tabfirst}97.1 & \cellcolor{tabfirst}0.070 & \cellcolor{tabfirst}96.5 \\
\cmidrule{2-9}
& Spann3R \cite{spann3r} & Stream & 0.622 & 42.6 & 0.144 & 81.3 & 0.198 & 73.7 \\
& Point3R \cite{wu2026point3r} & Stream & 0.452 & 48.9 & \cellcolor{tabsecond}0.060 & 96.0 & 0.136 & 84.2 \\
& CUT3R \cite{cut3r} & Stream & 0.421 & 47.9 & 0.078 & 93.7 & 0.118 & 88.1 \\
& TTT3R \cite{chenttt3r} & Stream & \cellcolor{tabthird}0.405 & 48.9 & 0.069 & 95.4 & \cellcolor{tabthird}0.114 & \cellcolor{tabthird}90.4 \\
& Stream3R \cite{lan2025stream3r} & Stream & 0.478 & \cellcolor{tabsecond}51.1 & 0.075 & 94.1 & 0.116 & 89.6 \\
& MUT3R \cite{shen2025mut3r} & Stream & 0.451 & 48.6 & 0.070 & \cellcolor{tabthird}96.2 & 0.116 & 88.3 \\
& TTSA3R \cite{zheng2026ttsa3r} & Stream & \cellcolor{tabsecond}0.401 & \cellcolor{tabthird}50.0 & \cellcolor{tabthird}0.064 & \cellcolor{tabsecond}96.5 & \cellcolor{tabsecond}0.110 & \cellcolor{tabsecond}91.2 \\
& Ours & Stream & \cellcolor{tabfirst}0.395 & \cellcolor{tabfirst}52.3 & \cellcolor{tabfirst}0.059 & \cellcolor{tabfirst}97.2 & \cellcolor{tabfirst}0.107 & \cellcolor{tabfirst}91.6 \\
\midrule
\multirow{6}{*}{Metric scale}
& MASt3R-GA \cite{mast3r} & Optim & 1.022 & 14.3 & 0.272 & 70.6 & 0.467 & 15.2 \\
& CUT3R \cite{cut3r} & Stream & 1.029 & 23.8 & 0.103 & 88.5 & 0.122 & 85.5 \\
& Point3R \cite{wu2026point3r} & Stream & \cellcolor{tabfirst}0.777 & 17.1 & 0.137 & 94.7 & 0.191 & 73.8 \\
& TTT3R \cite{chenttt3r} & Stream & 0.977 & 24.5 & 0.090 & 94.2 & \cellcolor{tabsecond}0.110 & \cellcolor{tabsecond}89.1 \\
& Stream3R \cite{lan2025stream3r} & Stream & 1.041 & 21.0 & \cellcolor{tabthird}0.084 & 94.4 & 0.234 & 57.6 \\
& MUT3R \cite{shen2025mut3r} & Stream & \cellcolor{tabsecond}0.820 & \cellcolor{tabsecond}25.2 & 0.086 & \cellcolor{tabthird}96.0 & 0.125 & 85.8 \\
& TTSA3R \cite{zheng2026ttsa3r} & Stream & 0.960 & \cellcolor{tabthird}24.6 & \cellcolor{tabsecond}0.079 & \cellcolor{tabsecond}96.6 & \cellcolor{tabthird}0.111 & \cellcolor{tabthird}88.9 \\
& Ours & Stream & \cellcolor{tabthird}0.945 & \cellcolor{tabfirst}25.6 & \cellcolor{tabfirst}0.075 & \cellcolor{tabfirst}96.8 & \cellcolor{tabfirst}0.108 & \cellcolor{tabfirst}89.4 \\
\bottomrule
\end{tabular}
\caption{Video Depth Estimation on standard short sequences. We evaluate scale-invariant and metric depth accuracy on Sintel\cite{sintel}, Bonn \cite{bonn}, and KITTI \cite{kitti} datasets.
Methods that require global alignment are denoted as "GA". "Optim", "Stream", and
"FA" refer to global-alignment optimization-based, streaming, and full-attention methods, respectively. Best results are highlighted as \colorbox{tabfirst}{first}, \colorbox{tabsecond}{second}, and \colorbox{tabthird}{third}.}
\label{tab:depth}
\end{table*}

Under the per-sequence scale setting, full-attention models achieve strong performance by jointly reasoning over all input frames, but their computational and memory costs increase with sequence length. Among streaming methods, our method achieves the best overall performance across all three datasets. It provides consistently strong Abs Rel and $\delta<1.25$ results on Sintel and Bonn and achieves the best streaming performance on KITTI. These results show that the proposed adaptive world memory preserves the dense geometric prediction ability of 3D foundation models while enabling online streaming inference.

The advantage becomes more evident under the metric-scale setting, where no per-sequence scale correction is applied. Our method improves over TTSA3R on all three datasets in terms of Abs Rel and achieves the best $\delta<1.25$ performance across the three benchmarks among streaming methods. Table~\ref{tab:depth_replica} further confirms strong long-sequence metric-depth performance on Replica. These improvements indicate that the dual-stage memory regulation better preserves metric consistency across frames by combining learned recurrent propagation with test-time temporal--spatial update control.

Overall, the depth results demonstrate that our method effectively balances long-term memory preservation and new observation integration. Compared with uniform or weakly adaptive recurrent updates, the proposed memory mechanism reduces geometric degradation during streaming inference while remaining competitive with full-attention models that require full-sequence processing.

\begin{table*}[!tbp]
		\centering
		\begin{tabular}{lllccccccccc}
		\toprule
		& & & office0 & office1 & office2 & office3 & office4 & room0 & room1 & room2 & average \\
		\midrule
		\multirow{14}{*}{Abs Rel $\downarrow$} & \multirow{2}{*}{CUT3R\cite{cut3r}} & Per-seq & 0.0419 & 0.2143 & 0.0384 & 0.0219 & 0.0362 & 0.0232 & 0.0422 & 0.0325 & 0.0564 \\
		& & Metric & 0.0432 & 0.2171 & 0.0746 & 0.032 & 0.0374 & 0.0361 & 0.0596 & 0.0564 & 0.0696 \\
		\cmidrule(lr){2-12}
		& \multirow{2}{*}{Point3R\cite{wu2026point3r}} & Per-seq & 0.105 & \cellcolor{tabsecond}0.1575 & 0.0684 & 0.05 & 0.0613 & 0.0498 & 0.0847 & 0.0748 & 0.0815 \\
		& & Metric & 0.1075 & \cellcolor{tabfirst}0.1714 & 0.0746 & 0.109 & 0.1952 & 0.0704 & 0.1687 & 0.0975 & 0.1244 \\
		\cmidrule(lr){2-12}
		& \multirow{2}{*}{TTT3R\cite{chenttt3r}} & Per-seq & 0.0393 & 0.1972 & 0.0342 & \cellcolor{tabsecond}0.0217 & \cellcolor{tabsecond}0.0255 & 0.0222 & 0.0412 & 0.0273 & 0.0512 \\
		& & Metric & 0.0391 & 0.2029 & \cellcolor{tabfirst}0.0643 & \cellcolor{tabfirst}0.0232 & 0.0342 & \cellcolor{tabsecond}0.0267 & \cellcolor{tabfirst}0.0509 & \cellcolor{tabfirst}0.0481 & \cellcolor{tabsecond}0.0612 \\
		\cmidrule(lr){2-12}
		& \multirow{2}{*}{Stream3R\cite{lan2025stream3r}} & Per-seq & \cellcolor{tabfirst}0.0335 & \cellcolor{tabfirst}0.0475 & \cellcolor{tabfirst}0.0264 & \cellcolor{tabfirst}0.0193 & \cellcolor{tabfirst}0.025 & \cellcolor{tabfirst}0.0195 & 0.0582 & \cellcolor{tabfirst}0.0194 & \cellcolor{tabfirst}0.0311 \\
		& & Metric & 0.6208 & 0.6166 & 0.7331 & 0.7303 & 0.7611 & 0.7143 & 0.6529 & 0.6726 & 0.6875 \\
		\cmidrule(lr){2-12}
		& \multirow{2}{*}{TTSA3R\cite{zheng2026ttsa3r}} & Per-seq & 0.0367 & 0.1934 & 0.0343 & \cellcolor{tabsecond}0.0217 & 0.0277 & 0.0219 & \cellcolor{tabfirst}0.0386 & 0.0261 & 0.0501 \\
		& & Metric & \cellcolor{tabsecond}0.0369 & 0.1974 & 0.0728 & 0.0283 & \cellcolor{tabsecond}0.0302 & 0.0269 & 0.0541 & 0.0513 & 0.0623 \\
		\cmidrule(lr){2-12}
		& \multirow{2}{*}{Ours} & Per-seq & \cellcolor{tabsecond}0.0355 & 0.1900 & \cellcolor{tabsecond}0.0329 & 0.0222 & 0.0275 & \cellcolor{tabsecond}0.0215 & \cellcolor{tabsecond}0.0402 & \cellcolor{tabsecond}0.0256 & \cellcolor{tabsecond}0.0494 \\
		& & Metric & \cellcolor{tabfirst}0.0361 & \cellcolor{tabsecond}0.196 & \cellcolor{tabsecond}0.0712 & \cellcolor{tabsecond}0.0274 & \cellcolor{tabfirst}0.0287 & \cellcolor{tabfirst}0.0255 & \cellcolor{tabsecond}0.0528 & \cellcolor{tabsecond}0.0504 & \cellcolor{tabfirst}0.0610 \\
		\midrule
		\multirow{14}{*}{$\delta<1.25$ $\uparrow$} & \multirow{2}{*}{CUT3R\cite{cut3r}} & Per-seq & \cellcolor{tabsecond}0.9958 & 0.4645 & 0.9883 & 0.9944 & 0.9924 & \cellcolor{tabsecond}0.997 & 0.9968 & 0.9938 & 0.9276 \\
		& & Metric & 0.9953 & 0.4889 & 0.9792 & \cellcolor{tabfirst}0.9944 & 0.9932 & \cellcolor{tabsecond}0.9969 & \cellcolor{tabfirst}0.9973 & 0.9926 & 0.9295 \\
		\cmidrule(lr){2-12}
		& \multirow{2}{*}{Point3R\cite{wu2026point3r}} & Per-seq & 0.8746 & \cellcolor{tabsecond}0.7053 & 0.9866 & 0.9937 & 0.9702 & 0.9802 & 0.9127 & 0.9369 & 0.9198 \\
		& & Metric & 0.8717 & \cellcolor{tabfirst}0.6887 & 0.9866 & 0.9685 & 0.5749 & 0.9896 & 0.7545 & 0.923 & 0.8442 \\
		\cmidrule(lr){2-12}
		& \multirow{2}{*}{TTT3R\cite{chenttt3r}} & Per-seq & 0.9953 & 0.5436 & 0.993 & 0.9944 & 0.9948 & 0.9969 & \cellcolor{tabsecond}0.9971 & 0.9941 & 0.9385 \\
		& & Metric & \cellcolor{tabfirst}0.9957 & 0.5163 & 0.9898 & 0.9943 & \cellcolor{tabfirst}0.995 & \cellcolor{tabsecond}0.9969 & \cellcolor{tabsecond}0.9969 & \cellcolor{tabsecond}0.9940 & 0.9346 \\
		\cmidrule(lr){2-12}
		& \multirow{2}{*}{Stream3R\cite{lan2025stream3r}} & Per-seq & \cellcolor{tabsecond}0.9956 & \cellcolor{tabfirst}0.9984 & \cellcolor{tabfirst}0.9974 & \cellcolor{tabfirst}0.9969 & \cellcolor{tabfirst}0.9973 & \cellcolor{tabfirst}0.9974 & 0.987 & \cellcolor{tabfirst}0.9963 & \cellcolor{tabfirst}0.9958 \\
		& & Metric & - & - & - & - & - & - & - & - & - \\
		\cmidrule(lr){2-12}
		& \multirow{2}{*}{TTSA3R\cite{zheng2026ttsa3r}} & Per-seq & 0.9952 & 0.5697 & 0.9941 & 0.9942 & \cellcolor{tabsecond}0.9949 & \cellcolor{tabsecond}0.9970 & \cellcolor{tabfirst}0.9974 & 0.9943 & 0.9419 \\
		& & Metric & 0.9951 & 0.5172 & \cellcolor{tabsecond}0.9899 & 0.994 & \cellcolor{tabfirst}0.995 & \cellcolor{tabsecond}0.9969 & \cellcolor{tabsecond}0.9969 & \cellcolor{tabsecond}0.9940 & \cellcolor{tabsecond}0.9347 \\
		\cmidrule(lr){2-12}
		& \multirow{2}{*}{Ours} & Per-seq & \cellcolor{tabfirst}0.9961 & 0.5704 & \cellcolor{tabsecond}0.9945 & \cellcolor{tabsecond}0.9947 & \cellcolor{tabsecond}0.9949 & \cellcolor{tabsecond}0.9970 & 0.9969 & \cellcolor{tabsecond}0.9952 & \cellcolor{tabsecond}0.9424 \\
		& & Metric & \cellcolor{tabsecond}0.9956 & \cellcolor{tabsecond}0.5263 & \cellcolor{tabfirst}0.9908 & \cellcolor{tabfirst}0.9944 & \cellcolor{tabfirst}0.9950 & \cellcolor{tabfirst}0.9970 & \cellcolor{tabsecond}0.9969 & \cellcolor{tabfirst}0.9942 & \cellcolor{tabfirst}0.9363 \\
		\bottomrule
		\end{tabular}
		\caption{Per-sequence scale and metric scale depth evaluation results on the Replica dataset\cite{replica}. The - symbol indicates that Stream3R does not support metric-scale depth evaluation. Best results are highlighted as \colorbox{tabfirst}{first}, and \colorbox{tabsecond}{second}.}
		\label{tab:depth_replica}
		\end{table*}

\subsection{Camera Pose Estimation}

We evaluate camera pose estimation on standard SLAM benchmarks, dynamic scenes, and long real-world sequences. For standard indoor scenes, Tables~\ref{tab:pose_7scenes} and~\ref{tab:pose_tum} show that our method achieves the best average trajectory accuracy among uncalibrated approaches on 7-Scenes and TUM RGB-D. The improvement mainly comes from the adaptive world memory, which preserves historical geometric information, and submap-level optimization, which reduces accumulated drift across local maps.

\begin{table*}[!tbp]
\centering
\setlength{\tabcolsep}{6pt}
\begin{tabular}{l|lccccccc|c}
\toprule
& \multirow{2}{*}{Method} & \multicolumn{7}{c|}{Sequence} & \multirow{2}{*}{Avg} \\
& & chess & fire & heads & office & pumpkin & kitchen & stairs & \\
\midrule
\multirow{3}{*}{\rotatebox{90}{Calib.}}
& NICER-SLAM \cite{nicerslam} & \cellcolor{tabfirst}0.033 & 0.069 & 0.042 & 0.108 & 0.200 & \cellcolor{tabfirst}0.039 & 0.108 & 0.086 \\
& DROID-SLAM \cite{droidslam} & \cellcolor{tabsecond}0.036 & \cellcolor{tabsecond}0.027 & \cellcolor{tabsecond}0.025 & \cellcolor{tabfirst}0.066 & \cellcolor{tabsecond}0.127 & \cellcolor{tabsecond}0.040 & \cellcolor{tabsecond}0.026 & \cellcolor{tabsecond}0.049 \\
& MASt3R-SLAM \cite{mast3rslam} & 0.053 & \cellcolor{tabfirst}0.025 & \cellcolor{tabfirst}0.015 & \cellcolor{tabsecond}0.097 & \cellcolor{tabfirst}0.088 & 0.041 & \cellcolor{tabfirst}0.011 & \cellcolor{tabfirst}0.047 \\
\midrule
\multirow{4}{*}{\rotatebox{90}{Uncalib.}}
& DROID-SLAM* \cite{droidslam} & 0.047 & 0.038 & 0.034 & 0.136 & 0.166 & 0.080 & 0.044 & 0.078 \\
& MASt3R-SLAM* \cite{mast3rslam} & 0.063 & 0.046 & 0.029 & \cellcolor{tabfirst}0.103 & \cellcolor{tabfirst}0.114 & 0.074 & \cellcolor{tabfirst}0.032 & \cellcolor{tabsecond}0.066 \\
& VGGT-SLAM \cite{maggio2026vggt} & \cellcolor{tabsecond}0.037 & \cellcolor{tabsecond}0.026 & \cellcolor{tabfirst}0.018 & 0.104 & 0.133 & \cellcolor{tabsecond}0.061 & 0.093 & 0.067 \\
& GRS-SLAM3R \cite{shen2025grs} & 0.069 & 0.050 & 0.032 & 0.113 & \cellcolor{tabsecond}0.125 & 0.081 & \cellcolor{tabsecond}0.035 & 0.072 \\
& Ours & \cellcolor{tabfirst}0.036 & \cellcolor{tabfirst}0.005 & \cellcolor{tabfirst}0.018 & \cellcolor{tabfirst}0.103 & 0.133 & \cellcolor{tabfirst}0.060 & 0.093 & \cellcolor{tabfirst}0.064 \\
\bottomrule
\end{tabular}
\caption{RMSE of ATE on 7-Scenes dataset \cite{7scenes} (unit: m). The * symbol indicates that the baseline
is evaluated in the uncalibrated mode. Best results are highlighted as \colorbox{tabfirst}{first}, and \colorbox{tabsecond}{second}.}
\label{tab:pose_7scenes}
\end{table*}

\begin{table*}[!tbp]
\centering
\setlength{\tabcolsep}{6pt}
\begin{tabular}{l|lcccccccccc}
\toprule
& & 360 & desk & desk2 & floor & plant & room & rpy & teddy & xyz & avg \\
\midrule
\multirow{8}{*}{\rotatebox{90}{Calibrated}}
& ORB-SLAM3 \cite{orbslam3} & X & \cellcolor{tabsecond}0.017 & 0.210 & X & 0.034 & X & X & X & \cellcolor{tabfirst}0.009 & - \\
& DeepV2D \cite{teed2018deepv2d} & 0.243 & 0.166 & 0.379 & 1.653 & 0.203 & 0.246 & 0.105 & 0.316 & 0.064 & 0.375 \\
& DeepFactors \cite{deepfactor} & 0.159 & 0.170 & 0.253 & 0.169 & 0.305 & 0.364 & 0.043 & 0.601 & 0.035 & 0.233 \\
& DPV-SLAM \cite{lipson2024deep} & 0.112 & 0.018 & \cellcolor{tabthird}0.029 & 0.057 & \cellcolor{tabthird}0.021 & 0.330 & 0.030 & 0.084 & \cellcolor{tabsecond}0.010 & 0.076 \\
& DPV-SLAM++ \cite{lipson2024deep} & 0.132 & 0.018 & \cellcolor{tabthird}0.029 & 0.050 & 0.022 & 0.096 & 0.032 & 0.098 & \cellcolor{tabsecond}0.010 & 0.054 \\
& GO-SLAM \cite{goslam} & \cellcolor{tabsecond}0.089 & \cellcolor{tabfirst}0.016 & \cellcolor{tabsecond}0.028 & \cellcolor{tabsecond}0.025 & 0.026 & \cellcolor{tabsecond}0.052 & \cellcolor{tabfirst}0.019 & \cellcolor{tabsecond}0.048 & \cellcolor{tabsecond}0.010 & \cellcolor{tabsecond}0.035 \\
& DROID-SLAM \cite{droidslam} & \cellcolor{tabthird}0.111 & 0.018 & 0.042 & \cellcolor{tabfirst}0.021 & \cellcolor{tabfirst}0.016 & \cellcolor{tabfirst}0.049 & \cellcolor{tabsecond}0.026 & \cellcolor{tabsecond}0.048 & 0.012 & \cellcolor{tabthird}0.038 \\
& MASt3R-SLAM \cite{mast3rslam} & \cellcolor{tabfirst}0.049 & \cellcolor{tabfirst}0.016 & \cellcolor{tabfirst}0.024 & \cellcolor{tabsecond}0.025 & \cellcolor{tabsecond}0.020 & \cellcolor{tabthird}0.061 & \cellcolor{tabthird}0.027 & \cellcolor{tabfirst}0.041 & \cellcolor{tabfirst}0.009 & \cellcolor{tabfirst}0.030 \\
\midrule
\multirow{6}{*}{\rotatebox{90}{Uncalibrated}}
& DROID-SLAM \cite{droidslam} & 0.202 & 0.032 & 0.091 & \cellcolor{tabthird}0.064 & 0.045 & 0.918 & 0.056 & 0.045 & \cellcolor{tabfirst}0.012 & 0.158 \\
& MASt3R-SLAM \cite{mast3rslam} & \cellcolor{tabsecond}0.070 & 0.035 & 0.055 & \cellcolor{tabfirst}0.056 & 0.035 & 0.118 & 0.041 & 0.114 & 0.020 & 0.060 \\
& ViSTA-SLAM \cite{zhang2026vista} & 0.104 & \cellcolor{tabthird}0.030 & \cellcolor{tabfirst}0.030 & 0.070 & 0.052 & \cellcolor{tabsecond}0.067 & \cellcolor{tabfirst}0.023 & 0.080 & 0.015 & \cellcolor{tabsecond}0.052 \\
& VGGT-SLAM Sim(3) \cite{maggio2026vggt} & 0.123 & 0.040 & 0.055 & 0.254 & \cellcolor{tabfirst}0.022 & \cellcolor{tabthird}0.088 & 0.041 & \cellcolor{tabfirst}0.032 & 0.016 & 0.074 \\
& VGGT-SLAM SL(4) \cite{maggio2026vggt} & \cellcolor{tabthird}0.071 & \cellcolor{tabsecond}0.025 & \cellcolor{tabthird}0.040 & 0.141 & \cellcolor{tabsecond}0.023 & 0.102 & \cellcolor{tabthird}0.030 & \cellcolor{tabsecond}0.034 & \cellcolor{tabsecond}0.014 & \cellcolor{tabthird}0.053 \\
& GRS-SLAM3R \cite{shen2025grs} & 0.076 & 0.038 & 0.060 & \cellcolor{tabsecond}0.061 & 0.038 & 0.129 & 0.045 & 0.125 & 0.022 & 0.066 \\

& Ours & \cellcolor{tabfirst}0.052 & \cellcolor{tabfirst}0.023 & \cellcolor{tabsecond}0.032 & 0.109 & \cellcolor{tabthird}0.024 & \cellcolor{tabfirst}0.060 & \cellcolor{tabsecond}0.026 & \cellcolor{tabthird}0.035 & \cellcolor{tabsecond}0.014 & \cellcolor{tabfirst}0.042 \\
\bottomrule
\end{tabular}
\caption{RMSE of ATE on TUM RGB-D \cite{tum} (unit: m). Best results are highlighted as \colorbox{tabfirst}{first}, \colorbox{tabsecond}{second}, and \colorbox{tabthird}{third}.}
\label{tab:pose_tum}
\end{table*}

We further evaluate pose robustness on Sintel, TUM-dynamics, and ScanNet in Table~\ref{tab:pose_dynamic}. Compared with streaming reconstruction methods using uniform or weakly adaptive state updates, our method obtains lower RMSE of ATE and more stable RPE on most benchmarks. This demonstrates that dual-stage memory regulation can suppress unreliable observations and prevent corrupted states from propagating into long-term memory, leading to more robust localization under dynamic content, rapid camera motion, and low-texture regions.

\begin{table*}[!tbp]
\centering
\setlength{\tabcolsep}{3pt}
\begin{tabular}{cc|ccc|ccc|ccc}
\toprule
\multirow{2}{*}{Method} & \multirow{2}{*}{Type}
& \multicolumn{3}{c|}{Sintel}
& \multicolumn{3}{c|}{TUM-dynamics}
& \multicolumn{3}{c}{ScanNet} \\
& & \shortstack{RMSE of\\ATE} $\downarrow$ & $RPE_{\mathrm{trans}}$ $\downarrow$ & $RPE_{\mathrm{rot}}$ $\downarrow$
& \shortstack{RMSE of\\ATE} $\downarrow$ & $RPE_{\mathrm{trans}}$ $\downarrow$ & $RPE_{\mathrm{rot}}$ $\downarrow$
& \shortstack{RMSE of\\ATE} $\downarrow$ & $RPE_{\mathrm{trans}}$ $\downarrow$ & $RPE_{\mathrm{rot}}$ $\downarrow$ \\
\midrule
Robust-CVD \cite{kopf2021robust} & Optim & 0.360 & 0.154 & 3.443 & 0.153 & 0.026 & 3.528 & 0.227 & 0.064 & 7.374 \\
CasualSAM \cite{zhang2022structure} & Optim & \cellcolor{tabthird}0.141 & \cellcolor{tabfirst}0.035 & \cellcolor{tabsecond}0.615 & \cellcolor{tabthird}0.071 & \cellcolor{tabfirst}0.010 & 1.712 & 0.158 & 0.034 & 1.618 \\
DUSt3R-GA \cite{dust3r} & Optim & 0.417 & 0.250 & 5.796 & 0.083 & 0.017 & 3.567 & 0.081 & 0.028 & 0.784 \\
MASt3R-GA \cite{mast3r} & Optim & 0.185 & 0.060 & 1.496 & \cellcolor{tabsecond}0.038 & \cellcolor{tabsecond}0.012 & \cellcolor{tabsecond}0.448 & 0.078 & 0.020 & \cellcolor{tabsecond}0.475 \\
MonST3R-GA \cite{zhang2025monst3r} & Optim & \cellcolor{tabsecond}0.111 & \cellcolor{tabthird}0.044 & 0.869 & 0.098 & 0.019 & \cellcolor{tabthird}0.935 & \cellcolor{tabthird}0.077 & \cellcolor{tabthird}0.018 & 0.529 \\
Easi3R \cite{chen2025easi3r} & FA & \cellcolor{tabfirst}0.110 & \cellcolor{tabsecond}0.042 & \cellcolor{tabthird}0.758 & 0.105 & 0.022 & 1.064 & \cellcolor{tabsecond}0.061 & \cellcolor{tabsecond}0.017 & \cellcolor{tabthird}0.525 \\
VGGT \cite{vggt} & FA & 0.172 & 0.062 & \cellcolor{tabfirst}0.471 & \cellcolor{tabfirst}0.012 & \cellcolor{tabfirst}0.010 & \cellcolor{tabfirst}0.310 & \cellcolor{tabfirst}0.035 & \cellcolor{tabfirst}0.015 & \cellcolor{tabfirst}0.377 \\
\midrule
Spann3R \cite{spann3r} & Stream & 0.329 & 0.110 & 4.471 & 0.056 & 0.021 & 0.591 & 0.096 & 0.023 & 0.661 \\
CUT3R \cite{cut3r} & Stream & 0.213 & \cellcolor{tabsecond}0.066 & \cellcolor{tabfirst}0.621 & 0.046 & 0.015 & 0.473 & 0.099 & 0.022 & \cellcolor{tabthird}0.600 \\
Point3R \cite{wu2026point3r} & Stream & 0.351 & 0.128 & 1.822 & 0.075 & 0.029 & 0.642 & 0.106 & 0.035 & 1.946 \\
TTT3R \cite{chenttt3r} & Stream & \cellcolor{tabsecond}0.210 & 0.090 & \cellcolor{tabsecond}0.722 & \cellcolor{tabsecond}0.028 & \cellcolor{tabsecond}0.013 & \cellcolor{tabsecond}0.380 & \cellcolor{tabthird}0.064 & \cellcolor{tabthird}0.021 & 0.637 \\
MUT3R \cite{shen2025mut3r} & Stream & 0.228 & \cellcolor{tabfirst}0.062 & 0.751 & 0.042 & 0.015 & 0.445 & - & - & - \\
TTSA3R \cite{zheng2026ttsa3r} & Stream & \cellcolor{tabsecond}0.210 & 0.085 & 0.765 & \cellcolor{tabfirst}0.026 & \cellcolor{tabfirst}0.012 & \cellcolor{tabfirst}0.372 & \cellcolor{tabsecond}0.057 & \cellcolor{tabsecond}0.020 & \cellcolor{tabsecond}0.588 \\
Ours & Stream & \cellcolor{tabfirst}0.207 & \cellcolor{tabthird}0.081 & \cellcolor{tabthird}0.750 & \cellcolor{tabthird}0.029 & \cellcolor{tabsecond}0.013 & \cellcolor{tabthird}0.385 & \cellcolor{tabfirst}0.050 & \cellcolor{tabfirst}0.019 & \cellcolor{tabfirst}0.518 \\
\bottomrule
\end{tabular}
\caption{Camera Pose Estimation on standard short sequences. We evaluate three
metrics on Sintel \cite{sintel}, TUM-dynamics \cite{tum}, and ScanNet \cite{scannet} datasets. RMSE of ATE and $RPE_{\mathrm{trans}}$ are reported in meters, while $RPE_{\mathrm{rot}}$ is reported in degrees. Optim, FA, and Stream denote optimization-based, full-attention, and streaming methods, respectively. Best results are highlighted as \colorbox{tabfirst}{first}, \colorbox{tabsecond}{second}, and \colorbox{tabthird}{third}.}
\label{tab:pose_dynamic}
\end{table*}

Additional long-sequence evaluations on KITTI and Replica are reported in Tables~\ref{tab:pose_kitti} and~\ref{tab:pose_replica}. On KITTI, our method achieves the best average among the listed learning-based baselines when the high-speed Sequence~01 is excluded. On Replica, it obtains the lowest average RMSE of ATE among the comparable 3D-foundation and uncalibrated baselines. Together, these results show that submap-aware memory management and global alignment maintain trajectory consistency over long sequences while preserving online operation.

\begin{table*}[!tbp]
			\centering
			\setlength{\tabcolsep}{4pt} 

			\begin{adjustbox}{width=\textwidth}
				\begin{tabular}{l|l|cc|cc|*{11}{c}}
				\toprule
				& & & & & & 00 & 01 & 02 & 03 & 04 & 05 & 06 & 07 & 08 & 09 & 10 \\
				\midrule
				Category & Methods & LC & Recon. & Avg. & Avg.* & & & & & & & & & & & \\
				\midrule
				\color{black!40} seq. frames & \color{black!40} - & \color{black!40} - & \color{black!40} - & \color{black!40} 2109 & \color{black!40} 2210 & \color{black!40} 4542 & \color{black!40} 1101 & \color{black!40} 4661 & \color{black!40} 801 & \color{black!40} 271 & \color{black!40} 2761 & \color{black!40} 1101 & \color{black!40} 1101 & \color{black!40} 4071 & \color{black!40} 1591 & \color{black!40} 1201 \\
				\color{black!40} seq. length (m) & \color{black!40} - & \color{black!40} - & \color{black!40} - & \color{black!40} 2012.243 & \color{black!40} 1968.147 & \color{black!40} 3724.19 & \color{black!40} 2453.20 & \color{black!40} 5067.23 & \color{black!40} 560.89 & \color{black!40} 393.65 & \color{black!40} 2205.58 & \color{black!40} 1232.88 & \color{black!40} 649.70 & \color{black!40} 3222.80 & \color{black!40} 1705.05 & \color{black!40} 919.52 \\
				\color{black!40} seq. speed (m/frame) & \color{black!40} - & \color{black!40} - & \color{black!40} - & \color{black!40} 0.95 & \color{black!40} 0.89 & \color{black!40} 0.82 & \color{black!40} 2.23 & \color{black!40} 1.09 & \color{black!40} 0.70 & \color{black!40} 1.45 & \color{black!40} 0.80 & \color{black!40} 1.12 & \color{black!40} 0.59 & \color{black!40} 0.79 & \color{black!40} 1.07 & \color{black!40} 0.77 \\
				\color{black!40} contains loop & \color{black!40} - & \color{black!40} - & \color{black!40} - & \color{black!40} - & \color{black!40} - & \color{black!40} $\checkmark$ & \color{black!40} $\times$ & \color{black!40} $\checkmark$ & \color{black!40} $\times$ & \color{black!40} $\times$ & \color{black!40} $\checkmark$ & \color{black!40} $\checkmark$ & \color{black!40} $\checkmark$ & \color{black!40} $\times$ & \color{black!40} $\checkmark$ & \color{black!40} $\times$ \\
				\midrule
				\multirow{3}{*}{Classic} & ORB-SLAM2 (w/o LC) \cite{orbslam2} & $\times$ & Sparse & 69.727 & 26.480 & 40.65 & 502.20 & 47.82 & 0.94 & 1.30 & 29.95 & 40.82 & 16.04 & 43.09 & 38.77 & 5.42 \\
				& ORB-SLAM2 (w/ LC) \cite{orbslam2} & $\checkmark$ & Sparse & 54.816 & 9.464 & 6.03 & 508.34 & 14.76 & 1.02 & 1.57 & 4.04 & 11.16 & 2.19 & 38.85 & 8.39 & 6.63 \\
				& LDSO \cite{gao2018ldso} & $\checkmark$ & Sparse & 22.425 & 23.500 & 9.32 & 11.68 & 31.98 & 2.85 & 1.22 & 5.10 & 13.55 & 2.96 & 129.02 & 21.64 & 17.36 \\
				\midrule
                    \multirow{13}{*}{Learning Based} & DROID-VO \cite{droidslam} & $\times$ & Dense & 54.188 & \cellcolor{tabsecond}51.187 & 98.43 & 84.20 & 108.80 & 2.58 & 0.93 & 59.27 & 64.40 & 24.20 & \cellcolor{tabfirst}64.55 & \cellcolor{tabsecond}71.80 & 16.91 \\
					& DPVO \cite{dpvo} & $\times$ & Sparse & 53.609 & 57.701 & 113.21 & \cellcolor{tabsecond}12.69 & 123.40 & \cellcolor{tabfirst}2.09 & \cellcolor{tabfirst}0.68 & \cellcolor{tabsecond}58.96 & \cellcolor{tabsecond}54.78 & \cellcolor{tabsecond}19.26 & 115.90 & 75.10 & \cellcolor{tabfirst}13.63 \\
					& DROID-SLAM \cite{droidslam} & - & Dense & 100.278 & 75.846 & \cellcolor{tabsecond}92.10 & 344.60 & \cellcolor{tabfirst}107.61 & \cellcolor{tabsecond}2.38 & 1.00 & 118.50 & 62.47 & 21.78 & 161.60 & 72.32 & 118.70 \\
					& DPV-SLAM \cite{lipson2024deep} & $\checkmark$ & Sparse & \cellcolor{tabfirst}53.034 & 57.187 & 112.80 & \cellcolor{tabfirst}11.50 & 123.53 & 2.50 & \cellcolor{tabsecond}0.81 & \cellcolor{tabfirst}57.80 & 54.86 & \cellcolor{tabfirst}18.77 & 110.49 & 76.66 & \cellcolor{tabsecond}13.65 \\
                    
					\cmidrule{2-17}
					& MASt3R-SLAM \cite{mast3rslam}  & $\checkmark$ & Dense & / & / & \color{black!40} TL & \color{black!40} TL & \color{black!40} TL & \color{black!40} TL & \color{black!40} TL & \color{black!40} TL & \color{black!40} TL & \color{black!40} TL & \color{black!40} TL & \color{black!40} TL & \color{black!40} TL \\
					& CUT3R \cite{cut3r} & $\times$ & Dense & / & / & \color{black!40} OOM & \color{black!40} OOM & \color{black!40} OOM & 148.07 & 22.31 & \color{black!40} OOM & \color{black!40} OOM & \color{black!40} OOM & \color{black!40} OOM & \color{black!40} OOM & \color{black!40} OOM \\
					& Fast3R \cite{yang2025fast3r} & $\times$ & Dense & / & / & \color{black!40} OOM & \color{black!40} OOM & \color{black!40} OOM & \color{black!40} OOM & \color{black!40} OOM & \color{black!40} OOM & \color{black!40} OOM & \color{black!40} OOM & \color{black!40} OOM & \color{black!40} OOM & \color{black!40} OOM \\
					& VGGT \cite{deng2025vggt} & $\times$ & Dense & / & / & \color{black!40} OOM & \color{black!40} OOM & \color{black!40} OOM & \color{black!40} OOM & \color{black!40} OOM & \color{black!40} OOM & \color{black!40} OOM & \color{black!40} OOM & \color{black!40} OOM & \color{black!40} OOM & \color{black!40} OOM \\
					\cmidrule{2-17}
					& Ours & $\checkmark$ & Dense & \cellcolor{tabsecond}53.509 & \cellcolor{tabfirst}44.762 & \cellcolor{tabfirst}57.63 & 140.98 & \cellcolor{tabsecond}107.90 & 7.25 & 2.81 & 74.13 & \cellcolor{tabfirst}31.03 & 23.39 & \cellcolor{tabsecond}78.06 & \cellcolor{tabfirst}41.71 & 23.71 \\
					\bottomrule
			\end{tabular}
			\end{adjustbox}
		\caption{RMSE of ATE on the KITTI Dataset Odometry Track\cite{Geiger2012CVPR} (unit: m). [LC] denotes loop closure. Since the Seq. 01 is a high-speed sequence, its movement pattern is significantly different from those of other sequences. \text{[Avg.*]} shows the mean RMSE of ATE excluding Seq 01. [OOM] is short for CUDA Out-Of-Memory on a single RTX 4090. [TL] is short for Tracking Lost. We do not compare with those classic methods. Best results are highlighted as \colorbox{tabfirst}{first}, and \colorbox{tabsecond}{second}.}
			\label{tab:pose_kitti}
		\end{table*}

\begin{table*}[!tbp]
		\centering
		\begin{tabular}{lccccccccc}
		\toprule
		Method & Room 0 & Room 1 & Room 2 & Office 0 & Office 1 & Office 2 & Office 3 & Office 4 & Average \\
		\midrule
		DUSt3R \cite{dust3r} & 7.69 & 8.28 & 11.41 & 7.00 & \cellcolor{tabthird}4.82 & 5.16 & 5.32 & 8.19 & 7.23 \\
		DUSt3R (gt int)\cite{dust3r} & \cellcolor{tabthird}3.97 & \cellcolor{tabsecond}4.80 & 7.65 & 5.10 & \cellcolor{tabsecond}3.99 & \cellcolor{tabthird}3.84 & \cellcolor{tabsecond}2.95 & \cellcolor{tabthird}6.42 & \cellcolor{tabsecond}4.84 \\
		MASt3R \cite{mast3r}& \cellcolor{tabfirst}1.60 & \cellcolor{tabfirst}1.88 & \cellcolor{tabsecond}4.05 & \cellcolor{tabfirst}1.57 & 6.72 & \cellcolor{tabfirst}1.49 & 75.09 & \cellcolor{tabsecond}5.09 & 12.19 \\
		\color{black!40} GO-SLAM \cite{goslam}& \color{black!40}- & \color{black!40}- & \color{black!40}- & \color{black!40}- & \color{black!40}- & \color{black!40}- & \color{black!40}- & \color{black!40}- & \color{black!40}0.39 \\
		\color{black!40} DIM-SLAM *\cite{li2023dense} & \color{black!40}1.06 & \color{black!40}0.49 & \color{black!40}0.32 & \color{black!40}0.43 & \color{black!40}0.26 & \color{black!40}0.65 & \color{black!40}0.55 & \color{black!40}3.69 & \color{black!40}0.93 \\
		\color{black!40} DROID-SLAM \cite{droidslam}& \color{black!40}0.34 & \color{black!40}0.13 & \color{black!40}0.27 & \color{black!40}0.25 & \color{black!40}0.42 & \color{black!40}0.32 & \color{black!40}0.52 & \color{black!40}0.40 & \color{black!40}0.33 \\
		\color{black!40} DROID-SLAM * \cite{droidslam}& \color{black!40}0.58 & \color{black!40}0.38 & \color{black!40}0.38 & \color{black!40}1.06 & \color{black!40}0.40 & \color{black!40}0.70 & \color{black!40}0.53 & \color{black!40}1.33 & \color{black!40}0.70 \\
		Spann3R \cite{spann3r} & 30.14 & 39.39 & 32.48 & 37.65 & 27.06 & 45.02 & 57.11 & 42.68 & 38.94 \\
		SLAM3R\cite{slam3r} & 4.62 & 5.95 & 5.46 & 10.96 & 7.18 & 5.85 & 5.27 & 16.33 & 7.70 \\
		CUT3R \cite{cut3r}& 22.56 & 40.99 & 20.52 & 21.56 & 26.13 & 24.23 & 16.47 & 22.54 & 24.37 \\
		MASt3R-SLAM \cite{mast3rslam}& 4.84 & 5.15 & \cellcolor{tabthird}4.21 & \cellcolor{tabthird}4.64 & 4.96 & 4.96 & \cellcolor{tabthird}4.67 & 7.36 & \cellcolor{tabthird}5.10 \\
		GRS-SLAM3R \cite{shen2025grs}& 5.15 & 6.19 & 5.76 & \cellcolor{tabsecond}3.77 & 6.44 & 5.19 & 4.89 & 7.18 & 5.57 \\
		Ours & \cellcolor{tabsecond}2.85 & \cellcolor{tabthird}4.93 & \cellcolor{tabfirst}3.39 & 4.69 & \cellcolor{tabfirst}2.19 & \cellcolor{tabsecond}2.16 & \cellcolor{tabfirst}1.90 & \cellcolor{tabfirst}2.30 & \cellcolor{tabfirst}3.05 \\
		\bottomrule
		\end{tabular}
		\caption{RMSE of ATE on the Replica dataset\cite{replica} (unit: cm).. The * symbol indicates that the baseline is evaluated in the uncalibrated mode. Gray rows denote conventional methods reported for reference and are excluded from the ranking. Best results are highlighted as \colorbox{tabfirst}{first}, \colorbox{tabsecond}{second}, and \colorbox{tabthird}{third}.}
		\label{tab:pose_replica}
		\end{table*}

\subsection{Dense Reconstruction Evaluation}

We evaluate dense reconstruction on public indoor benchmarks, large multi-room scenes, and self-collected indoor, outdoor, and indoor--outdoor sequences. Tables~\ref{tab:reconstruction_7scenes} and~\ref{tab:reconstruction_nrgbd} report the quantitative results on 7-Scenes and NRGBD. Our method achieves the best average accuracy and completeness on 7-Scenes, together with the best mean accuracy and completeness on NRGBD. These results demonstrate that the proposed memory representation preserves accurate local surfaces while improving overall reconstruction completeness.

\begin{table*}[!tbp]
		\centering
		\begin{adjustbox}{width=\textwidth}
		\begin{tabular}{l *{8}{cc}} 
		\toprule
		\multirow{2}{*}{\textbf{Method}} & 
		\multicolumn{2}{c}{\textbf{Chess}} & 
		\multicolumn{2}{c}{\textbf{Fire}} & 
		\multicolumn{2}{c}{\textbf{Heads}} & 
		\multicolumn{2}{c}{\textbf{Office}} & 
		\multicolumn{2}{c}{\textbf{Pumpkin}} & 
		\multicolumn{2}{c}{\textbf{RedKitchen}} & 
		\multicolumn{2}{c}{\textbf{Stairs}} & 
		\multicolumn{2}{c}{\textbf{Average}} \\
		\cmidrule(r){2-3} \cmidrule(lr){4-5} \cmidrule(lr){6-7} \cmidrule(lr){8-9} 
		\cmidrule(lr){10-11} \cmidrule(lr){12-13} \cmidrule(lr){14-15} \cmidrule(l){16-17}
		& Acc. & Comp. & Acc. & Comp. & Acc. & Comp. & Acc. & Comp. & Acc. & Comp. & Acc. & Comp. & Acc. & Comp. & Acc. & Comp. \\
		\midrule
		DUSt3R\cite{dust3r}& 2.26 & 2.13 & \cellcolor{tabthird}1.04 & 1.50 & 1.66 & \cellcolor{tabthird}0.98 & 4.62 & 4.74 & \cellcolor{tabfirst}1.73 & 2.43 & \cellcolor{tabsecond}1.95 & 2.36 & 3.37 & 10.75 & 2.19 & 3.24 \\
		MASt3R \cite{mast3r}& 2.08 & 2.12 & 1.54 & 1.43 & \cellcolor{tabsecond}1.06 & 1.04 & \cellcolor{tabsecond}3.23 & 3.19 & 5.68 & 3.07 & 3.50 & 3.37 & \cellcolor{tabsecond}2.36 & 13.16 & 3.04 & 3.90 \\
		Spann3R \cite{spann3r}& 2.23 & 1.68 & \cellcolor{tabsecond}0.88 & \cellcolor{tabsecond}0.92 & 2.67 & \cellcolor{tabthird}0.98 & 5.86 & 3.54 & 2.25 & \cellcolor{tabfirst}1.85 & 2.68 & \cellcolor{tabsecond}1.80 & 5.65 & \cellcolor{tabthird}5.15 & 3.42 & 2.41 \\
		CUT3R\cite{cut3r} & 2.46 & 1.99 & 1.52 & 1.43 & 2.10 & 1.13 & 3.81 & 3.05 & 2.98 & 2.48 & 2.49 & 2.24 & 3.35 & 10.53 & 2.67 & 3.27 \\
		SLAM3R \cite{slam3r}& \cellcolor{tabthird}1.63 & \cellcolor{tabsecond}1.31 & \cellcolor{tabfirst}0.84 & \cellcolor{tabfirst}0.83 & 2.95 & 1.22 & \cellcolor{tabfirst}2.32 & \cellcolor{tabfirst}2.26 & \cellcolor{tabsecond}1.81 & \cellcolor{tabthird}2.05 & \cellcolor{tabfirst}1.84 & \cellcolor{tabthird}1.94 & 4.19 & 6.91 & \cellcolor{tabthird}2.13 & \cellcolor{tabthird}2.34 \\
		MASt3R-SLAM\cite{mast3rslam} & 2.41 & 1.70 & 1.57 & 1.33 & 1.71 & 1.16 & \cellcolor{tabthird}3.47 & \cellcolor{tabthird}2.98 & 2.86 & 2.37 & 2.83 & 2.16 & 3.32 & 9.53 & 2.60 & 3.03 \\
		GRS-SLAM3R \cite{shen2025grs}& \cellcolor{tabsecond}1.49 & \cellcolor{tabthird}1.32 & 1.26 & 1.32 & \cellcolor{tabthird}1.22 & \cellcolor{tabsecond}0.83 & 4.17 & 3.41 & 2.27 & 2.25 & \cellcolor{tabthird}2.19 & 2.19 & \cellcolor{tabfirst}2.22 & \cellcolor{tabsecond}4.55 & \cellcolor{tabsecond}2.12 & \cellcolor{tabsecond}2.27 \\
		Ours & \cellcolor{tabfirst}1.42 & \cellcolor{tabfirst}1.16 & 1.15 & \cellcolor{tabthird}1.07 & \cellcolor{tabfirst}0.62 & \cellcolor{tabfirst}0.49 & 3.56 & \cellcolor{tabsecond}2.85 & \cellcolor{tabthird}1.93 & \cellcolor{tabsecond}1.99 & 2.21 & \cellcolor{tabfirst}1.72 & \cellcolor{tabthird}2.89 & \cellcolor{tabfirst}2.95 & \cellcolor{tabfirst}1.97 & \cellcolor{tabfirst}1.75 \\
		\bottomrule
		\end{tabular}
		\end{adjustbox}
		\caption{Reconstruction results on 7-Scenes\cite{7scenes}. Best results are highlighted as \colorbox{tabfirst}{first}, \colorbox{tabsecond}{second}, and \colorbox{tabthird}{third}.}
		\label{tab:reconstruction_7scenes}
		\end{table*}

\begin{table}[!htbp]
		\centering
		\footnotesize
		\begin{tabular}{lcccc} 
		\toprule
		\textbf{Method} & \multicolumn{2}{c}{\textbf{Acc.[cm]}} & \multicolumn{2}{c}{\textbf{Comp.[cm]}} \\
		\cmidrule(r){2-3} \cmidrule(l){4-5}
		& Mean & Median & Mean & Median \\
		\midrule
		DUSt3R-GA \cite{dust3r}& 0.144 & \cellcolor{tabfirst}0.019 & 0.154 & \cellcolor{tabfirst}0.018 \\
		MASt3R-GA \cite{mast3r}& \cellcolor{tabsecond}0.085 & 0.033 & \cellcolor{tabfirst}0.063 & 0.028 \\
		Spann3R  \cite{spann3r} & 0.416 & 0.323 & 0.417 & 0.285 \\
		CUT3R   \cite{cut3r}  & 0.099 & \cellcolor{tabthird}0.031 & 0.076 & \cellcolor{tabthird}0.026 \\
		GRS-SLAM3R \cite{shen2025grs} & \cellcolor{tabthird}0.089 & \cellcolor{tabsecond}0.030 & \cellcolor{tabsecond}0.072 & \cellcolor{tabsecond}0.025 \\
		Ours      & \cellcolor{tabfirst}0.055 & 0.032 & \cellcolor{tabfirst}0.063 & 0.031 \\
		\bottomrule
		\end{tabular}
		\caption{Reconstruction results on the NRGBD dataset \cite{neural_rgbd}. Best results are highlighted as \colorbox{tabfirst}{first}, \colorbox{tabsecond}{second}, and \colorbox{tabthird}{third}.}
		\label{tab:reconstruction_nrgbd}
		\end{table}

\begin{figure*}[!tbp]
  \centering
  \includegraphics[width=\linewidth]{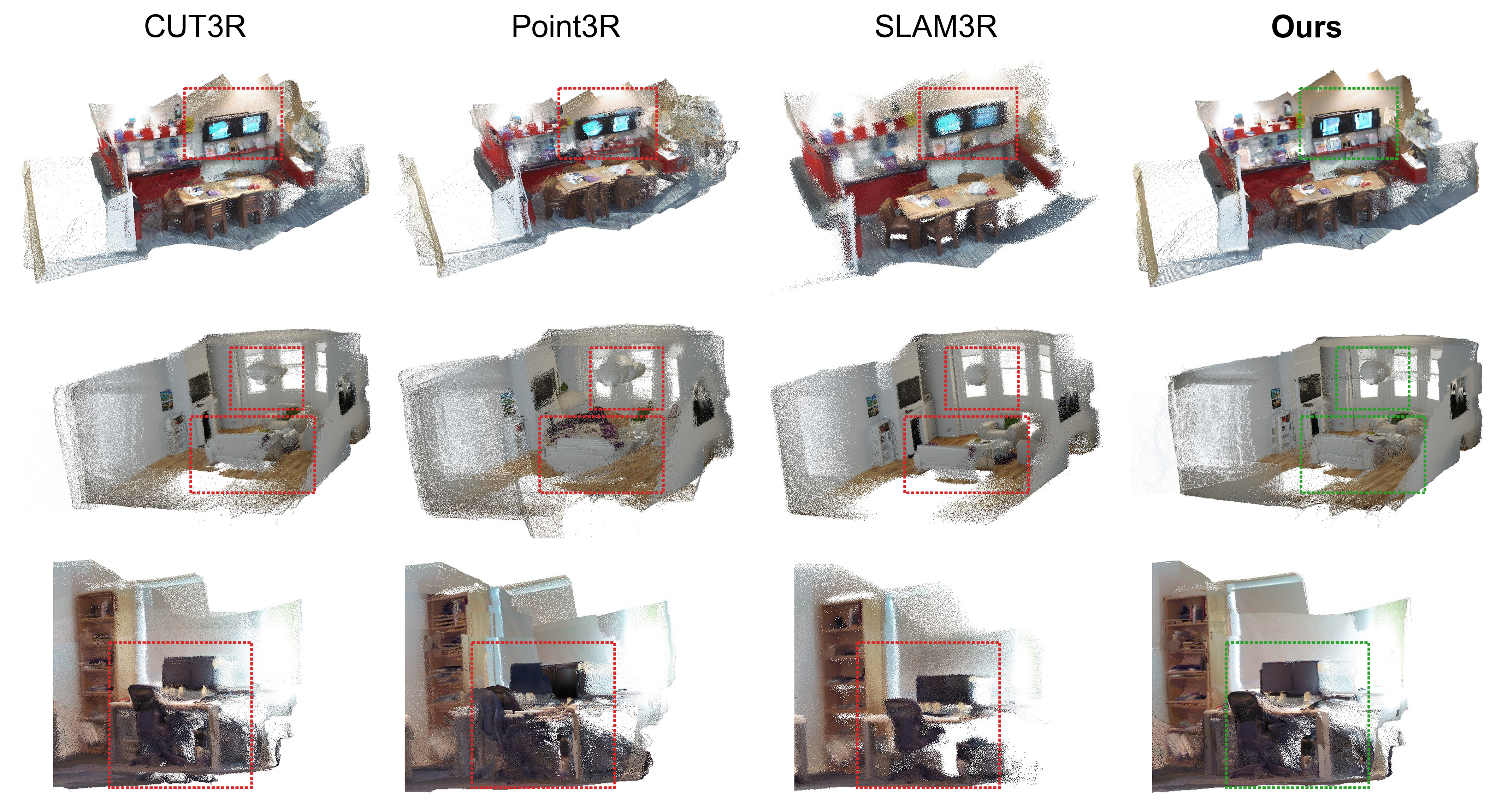}
  \caption{\textbf{Qualitative scene reconstruction results. } From left to right, we compare CUT3R\cite{cut3r}, Point3R\cite{wu2026point3r}, SLAM3R\cite{slam3r}, and our method on the redkitchen seq-06, whiteroom, and office seq-02 scenes (top to bottom) from 7-Scenes\cite{7scenes} and NRGBD\cite{neural_rgbd} datasets. Red dashed boxes highlight representative regions reconstructed by competing methods, while green dashed boxes indicate the corresponding regions reconstructed by our method. Our method recovers more complete geometry and preserves finer structural details.}
  \label{fig:reconstruction}
\end{figure*}

We further evaluate large-scale reconstruction on Replica/Apartment and the self-collected multi-platform dataset. Qualitative comparisons in Figs.~\ref{fig:reconstruction} and~\ref{fig:realworld} show that our method reconstructs more complete room structures, preserves finer geometric details, and produces globally coherent layouts in large indoor and indoor--outdoor environments. In particular, the indoor--outdoor sequence illustrates that the system can maintain consistent geometry across substantial appearance and structural changes.

The improvement is mainly attributed to the adaptive world memory and submap-aware map organization. The dual-stage update preserves reliable historical geometry while avoiding the accumulation of noisy observations, improving local surface quality. Meanwhile, submap-level alignment and loop-aware optimization reduce long-range drift and duplicated structures, improving global reconstruction consistency in multi-room and long-trajectory scenes.

\subsection{Gaussian Rendering Evaluation}

We evaluate the rendering quality of the proposed Gaussian world representation on ScanNet~\cite{scannet} and BundleFusion~\cite{Bundle-fusion}. 
These datasets contain indoor scenes with different spatial scales and camera trajectories, allowing us to assess whether the Gaussian reconstruction head can produce high-fidelity renderable maps from the shared adaptive world memory. 
We report PSNR, SSIM, and LPIPS on held-out views, as shown in Table~\ref{tab:rendering}.

\begin{table*}[!tbp]
		\centering
		\setlength{\tabcolsep}{5pt} 
		
		\begin{tabular}{llcccccc|cccccc}
			\toprule
			\multirow{2}{*}{Method} & \multirow{2}{*}{Metric} & \multicolumn{6}{c|}{ScanNet} & \multicolumn{6}{c}{BundleFusion} \\
			\cmidrule(l){3-8} \cmidrule(l){9-14}
			& & 0054 & 0059 & 0106 & 0169 & 0233 & 0465 & apt0 & apt2 & copyroom & office0 & office2 \\
			\midrule
			
			\multirow{3}{*}{MonoGS\cite{monogs}} 
			& SSIM $\uparrow$ & \cellcolor{tabsecond}0.80 & \cellcolor{tabsecond}0.74 & 0.72 & \cellcolor{tabsecond}0.77 & 0.68 & 0.59 & \cellcolor{tabsecond}0.70 & 0.39 & 0.70 & 0.52 & 0.60 \\
			& LPIPS $\downarrow$ & 0.61 & 0.60 & 0.54 & 0.66 & 0.67 & 0.74 & 0.67 & 0.82 & 0.63 & 0.78 & 0.71 \\
			& PSNR $\uparrow$ & 19.24 & 16.54 & 16.09 & \cellcolor{tabfirst}18.86 & 17.65 & 14.52 & 13.68 & 11.50 & 14.37 & 13.38 & 13.96 \\
			\midrule
			
			\multirow{3}{*}{DepthGS\cite{zhao2025pseudo}} 
			& SSIM $\uparrow$ & 0.31 & 0.32 & 0.34 & 0.42 & 0.36 & 0.26 & 0.38 & 0.41 & 0.58 & 0.56 & 0.58 \\
			& LPIPS $\downarrow$ & 0.79 & 0.78 & 0.78 & 0.73 & 0.84 & 0.81 & 0.67 & 0.69 & 0.51 & 0.62 & 0.63 \\
			& PSNR $\uparrow$ & 12.29 & 12.42 & 11.76 & 13.64 & 13.17 & 11.11 & 13.65 & 14.85 & 17.00 & 15.96 & 16.51 \\
			\midrule
			
			\multirow{3}{*}{S3PO-GS\cite{cheng2025outdoor}} 
			& SSIM $\uparrow$ & \cellcolor{tabsecond}0.80 & 0.71 & \cellcolor{tabsecond}0.75 & \cellcolor{tabfirst}0.78 & \cellcolor{tabfirst}0.73 & 0.61 & \cellcolor{tabfirst}0.74 & \cellcolor{tabsecond}0.64 & 0.47 & 0.63 & \cellcolor{tabsecond}0.64 \\
			& LPIPS $\downarrow$ & 0.62 & 0.58 & 0.54 & 0.55 & 0.69 & 0.75 & 0.57 & 0.71 & 0.78 & 0.71 & 0.64 \\
			& PSNR $\uparrow$ & 20.79 & 17.19 & 17.60 & \cellcolor{tabsecond}18.52 & 18.37 & 14.14 & 18.98 & 15.72 & 18.56 & 15.23 & 16.59 \\
			\midrule

			\multirow{3}{*}{Flash-Mono\cite{zhang2026flash}} 
			& SSIM $\uparrow$ & 0.79 & 0.66 & 0.72 & 0.73 & 0.69 & \cellcolor{tabsecond}0.66 & 0.66 & 0.60 & \cellcolor{tabsecond}0.72 & \cellcolor{tabsecond}0.69 & \cellcolor{tabsecond}0.64 \\
			& LPIPS $\downarrow$ & \cellcolor{tabsecond}0.39 & \cellcolor{tabsecond}0.41 & \cellcolor{tabsecond}0.43 & \cellcolor{tabsecond}0.39 & \cellcolor{tabsecond}0.44 & \cellcolor{tabsecond}0.45 & \cellcolor{tabsecond}0.49 & \cellcolor{tabsecond}0.54 & \cellcolor{tabsecond}0.45 & \cellcolor{tabsecond}0.50 & \cellcolor{tabsecond}0.51 \\
			& PSNR $\uparrow$ & \cellcolor{tabsecond}21.73 & \cellcolor{tabsecond}17.83 & \cellcolor{tabsecond}17.75 & \cellcolor{tabsecond}18.52 & \cellcolor{tabsecond}21.60 & \cellcolor{tabsecond}19.51 & \cellcolor{tabsecond}19.03 & \cellcolor{tabsecond}16.48 & \cellcolor{tabsecond}19.50 & \cellcolor{tabsecond}17.10 & \cellcolor{tabsecond}17.63 \\
			\midrule

			\multirow{3}{*}{Ours} 
			& SSIM $\uparrow$ & \cellcolor{tabfirst}0.82 & \cellcolor{tabfirst}0.80 & \cellcolor{tabfirst}0.78 & \cellcolor{tabsecond}0.77 & \cellcolor{tabsecond}0.72 & \cellcolor{tabfirst}0.83 & 0.68 & \cellcolor{tabfirst}0.73 & \cellcolor{tabfirst}0.80 & \cellcolor{tabfirst}0.72 & \cellcolor{tabfirst}0.73 \\
			& LPIPS $\downarrow$ & \cellcolor{tabfirst}0.26 & \cellcolor{tabfirst}0.25 & \cellcolor{tabfirst}0.33 & \cellcolor{tabfirst}0.32 & \cellcolor{tabfirst}0.31 & \cellcolor{tabfirst}0.23 & \cellcolor{tabfirst}0.40 & \cellcolor{tabfirst}0.41 & \cellcolor{tabfirst}0.35 & \cellcolor{tabfirst}0.40 & \cellcolor{tabfirst}0.37 \\
			& PSNR $\uparrow$ & \cellcolor{tabfirst}23.23 & \cellcolor{tabfirst}20.46 & \cellcolor{tabfirst}19.22 & 17.45 & \cellcolor{tabfirst}22.29 & \cellcolor{tabfirst}25.75 & \cellcolor{tabfirst}20.84 & \cellcolor{tabfirst}22.96 & \cellcolor{tabfirst}23.83 & \cellcolor{tabfirst}20.13 & \cellcolor{tabfirst}20.57 \\
			
			\bottomrule
		\end{tabular}
		\caption{Comparison rendering results on ScanNet \cite{scannet} and BundleFusion\cite{Bundle-fusion} datasets. Best results are highlighted as \colorbox{tabfirst}{first}, and \colorbox{tabsecond}{second}.}
		\label{tab:rendering}
		\end{table*}

Compared with existing neural rendering and Gaussian SLAM baselines, our method achieves superior rendering quality while jointly maintaining accurate camera tracking and dense reconstruction. As reported in Table~\ref{tab:rendering}, it obtains the lowest LPIPS on every evaluated scene, the highest PSNR on nearly all scenes, and the highest SSIM on most scenes. The improvement mainly comes from memory-conditioned Gaussian prediction: metric pointmaps provide reliable geometric anchors, while the Gaussian reconstruction head predicts appearance-aware primitives from memory-enhanced features.

The rendering results show that our method produces sharper textures, fewer floating artifacts, and more consistent appearance across viewpoints, especially in multi-room and long-sequence scenes. These results demonstrate that the adaptive world memory can be decoded not only into geometric maps, but also into a photorealistic and renderable scene representation.

\subsection{Real-World Multi-Platform Evaluation}

We further evaluate our method on self-collected real-world sequences captured by different robotic platforms, including handheld devices, wheeled mobile robots, and quadruped robots. 
As summarized in Table~\ref{tab:dataset_summary}, these platforms introduce diverse motion patterns and sensing conditions, ranging from smooth handheld motion to wheel-induced planar motion and legged locomotion with stronger vibration. 
The collected scenes cover office rooms, long corridors, multi-room indoor environments, and indoor--outdoor hybrid scenarios in both Singapore and Shanghai.

Since accurate ground-truth poses and dense geometry are difficult to obtain for every large-scale robot sequence, we combine quantitative trajectory evaluation with qualitative reconstruction and rendering comparisons. Table~\ref{tab:realworld_pose} reports trajectory errors on five representative sequences. Our method achieves the best average RMSE of ATE among the compared approaches, ranks first on three sequences, and ranks second on the remaining two, demonstrating robust localization across handheld, wheeled, and quadruped platforms.

\begin{table*}[!tbp]
		\centering
		\begin{tabular}{lcccccc}
		\toprule
		& e\_valley & 216 & 2F\_long\_loop & mid\_indoor\_mono & mid\_indoor\_mono\_2 & average \\ \midrule
		
		SLAM3R\cite{slam3r} & 2.15 & \cellcolor{tabthird}1.06 & 12.43 & \cellcolor{tabthird}0.25 & 0.32 & 3.24 \\
		MASt3R-SLAM\cite{mast3rslam} & \cellcolor{tabsecond}0.84 & \cellcolor{tabfirst}0.30 & \cellcolor{tabfirst}1.16 & 0.27 & 0.26 & \cellcolor{tabsecond}0.57 \\
		DROID-SLAM \cite{droidslam}& 3.50 & 2.63 & 8.04 & 0.39 & \cellcolor{tabsecond}0.085 & 2.93 \\
		DPVO \cite{dpvo}& 3.48 & 1.95  & \cellcolor{tabthird}1.28 & 0.61 & \cellcolor{tabthird}0.22 & \cellcolor{tabthird}1.51 \\
        GRS-SLAM3R~\cite{shen2025grs} & \cellcolor{tabthird}1.81 & 1.11 & 11.44 & \cellcolor{tabsecond}0.17 & \cellcolor{tabthird}0.22 & 2.95 \\
		Ours & \cellcolor{tabfirst}0.44 & \cellcolor{tabsecond}0.75 & \cellcolor{tabsecond}1.19 & \cellcolor{tabfirst}0.07 & \cellcolor{tabfirst}0.06 & \cellcolor{tabfirst}0.50 \\
		\bottomrule
		\end{tabular}
        \caption{Comparison of RMSE of ATE on our dataset (unit: m). Best results are highlighted as \colorbox{tabfirst}{first}, \colorbox{tabsecond}{second}, and \colorbox{tabthird}{third}.}
		\label{tab:realworld_pose}
		\end{table*}

\begin{figure*}[!tbp]
  \centering
  \includegraphics[width=\linewidth]{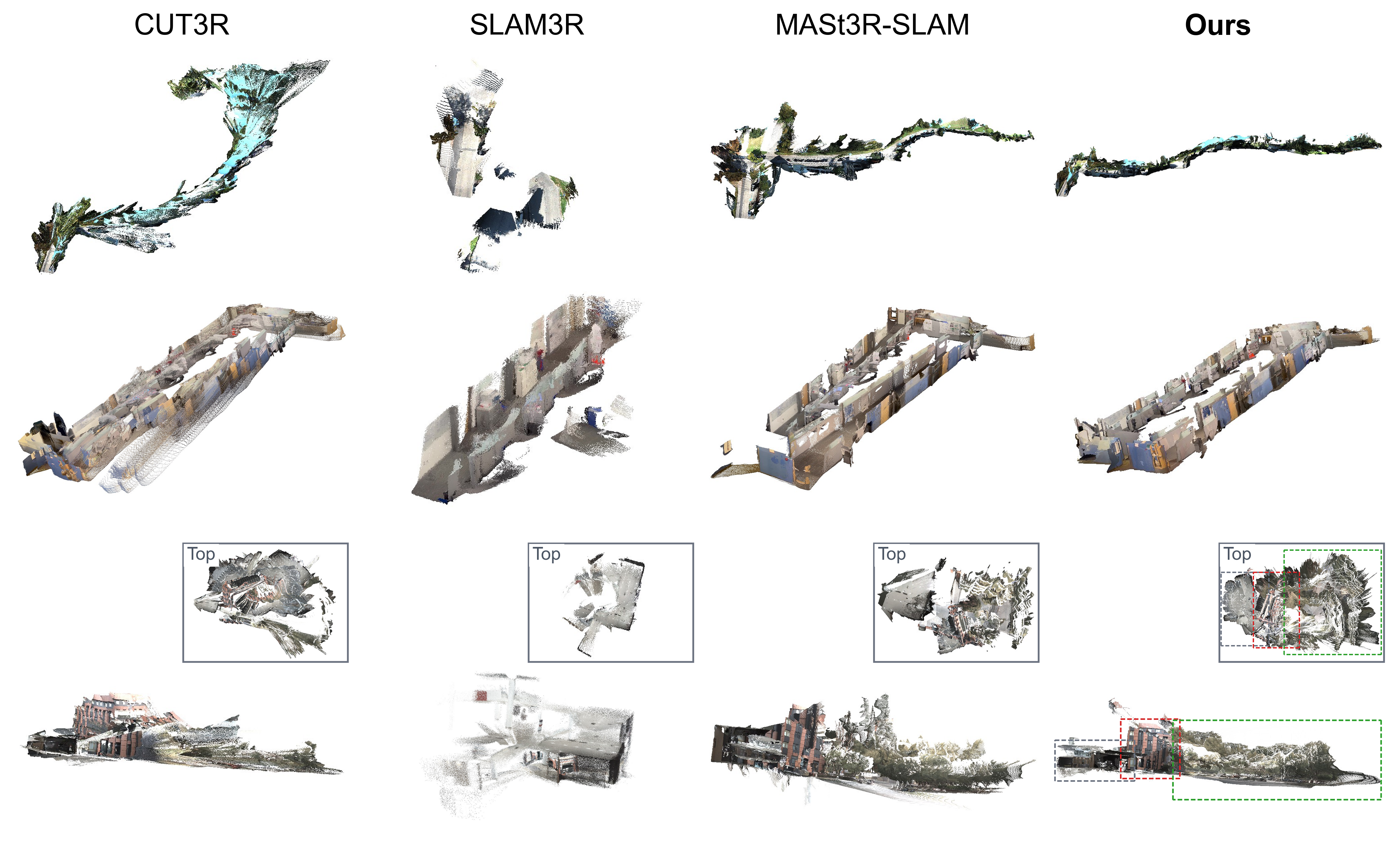}
  \caption{\textbf{Qualitative scene reconstruction results.} From left to right, we compare CUT3R\cite{cut3r}, SLAM3R\cite{slam3r}, MASt3R-SLAM\cite{mast3rslam}, and our method on the KITTI Odometry sequence 03\cite{Geiger2012CVPR}, office\_loop, and mid\_fir\_floor\_mono scenes (top to bottom). For the mid\_fir\_floor\_mono scene, the top-view insets facilitate comparison of the global scene layout. In the Ours column, the gray, red, and green dashed boxes denote the indoor area, building exterior, and outdoor area, respectively, in both the top and perspective views. Our method produces more complete and globally coherent reconstructions of large-scale environments.}
  \label{fig:realworld}
\end{figure*}

Fig.~\ref{fig:realworld} shows that our method produces more complete and globally coherent reconstructions on KITTI Odometry~03, office\_loop, and mid\_fir\_floor\_mono. In long corridors and multi-room environments, submap-aware memory prevents excessive memory growth and maintains stable local reconstruction. In the indoor--outdoor sequence, adaptive memory regulation reduces the influence of illumination changes, motion blur, and unreliable observations while preserving global scene structure. The corresponding rendering results further show that the shared memory can produce coherent Gaussian representations across different platforms and environments.

Compared with standard indoor benchmarks, these sequences contain stronger viewpoint changes, platform-dependent motion disturbances, repetitive structures, dynamic objects, and illumination variation. The results demonstrate that the proposed memory-centric 3D foundation model generalizes beyond controlled RGB-D benchmarks and can support localization, dense reconstruction, and renderable mapping in practical robotic scenarios.

\subsection{Runtime and Memory Analysis}

\begin{figure}[!htbp]
  \centering
  \includegraphics[width=0.95\linewidth]{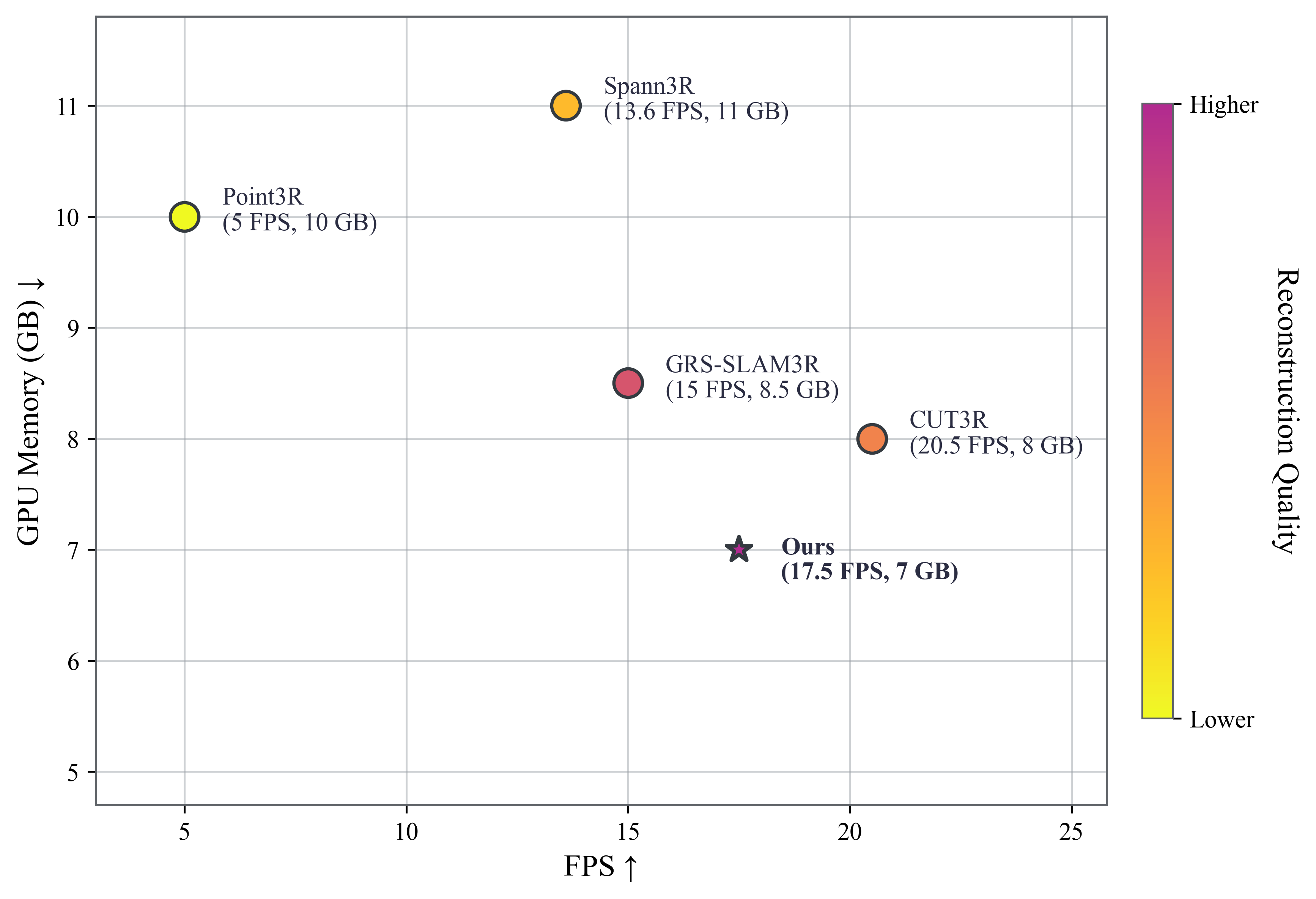}
  \caption{\textbf{Comparison of inference efficiency and reconstruction quality.} We compare the inference speed (FPS) and GPU memory consumption of different methods under the same evaluation setting. Marker colors follow the adjacent orange-yellow-to-red-purple scale from lower to higher reconstruction quality, and each annotation gives the corresponding FPS and memory usage. Our method achieves the best reconstruction quality while using only 7 GB of GPU memory and running at 17.5 FPS.}
  \label{fig:infer_eff}
\end{figure}

We analyze the runtime efficiency and GPU memory consumption of the proposed method during online inference in Fig.~\ref{fig:infer_eff}. All experiments are conducted on a single NVIDIA RTX~4090 GPU. We report end-to-end throughput and the average processing time of the main system components, including feature extraction, adaptive memory update, pose tracking, Gaussian map update, rendering, and submap optimization. The proposed method processes frames sequentially and does not require full-sequence attention, making it suitable for long video streams and online robotic operation.

Compared with full-attention 3D foundation models, our method significantly reduces memory growth with respect to sequence length by maintaining a fixed-size adaptive world memory and organizing the scene into local submaps. The memory update introduces only a small computational overhead while improving long-range consistency and reducing tracking drift. The Gaussian reconstruction head enables real-time renderable map construction, and submap optimization is triggered only when new submaps or loop-closure constraints are detected, avoiding unnecessary global optimization at every frame.

The results show that our method achieves a favorable balance among reconstruction quality, localization accuracy, rendering quality, and online efficiency. In particular, the submap-aware design keeps GPU memory bounded over long sequences, while the progressive mapping and tracking module remains efficient together with Gaussian rendering for practical robotic deployment.

\subsection{Ablation Studies}

We conduct ablation studies to analyze the contribution of gated recurrent memory, test-time temporal--spatial regulation, submap-aware memory management, SL(4)-based global optimization, and the Gaussian reconstruction head. Table~\ref{tab:ablation} separates the original online regulation into temporal (Temp.) and spatial (Spat.) components on the Apartment dataset\cite{replica}. The temporal-only and spatial-only settings evaluate the two cues independently, while enabling both components corresponds to temporal--spatial fused regulation. Removing the gated state degrades pose estimation and reconstruction quality, while removing online temporal--spatial regulation increases trajectory drift, especially under dynamic or low-quality observations. Removing submap organization leads to the largest performance degradation, demonstrating its importance for long-sequence memory management.

\begin{table}[!htbp]
		\centering
		\resizebox{\columnwidth}{!}{%
		\begin{tabular}{ccccccccc}
		\toprule
		\multicolumn{6}{c}{Modules} & \multicolumn{2}{c}{Reconstruction (cm)} & \multirow{2}{*}{\shortstack{RMSE of\\ATE (m)}} \\
		\cmidrule(r){1-6} \cmidrule(r){7-8}
		Gate & Temp. & Spat. & Submap & LC & W & Acc. & Comp. & \\
		\midrule
		\xmark & \xmark & \xmark & \xmark & \xmark & - & 76.53 & 59.70 & 2.98 \\
		\xmark & \cmark & \cmark & \cmark & \xmark & 16 & 16.31 & 13.74 & 0.55\\
		\cmark & \cmark & \cmark & \xmark & \xmark & - & 35.19 & 29.37 & 1.42\\
		\cmark & \xmark & \xmark & \cmark & \xmark & 16 & 20.40 & 17.39 & 0.70\\
		\cmark & \cmark & \xmark & \cmark & \xmark & 16 & 13.26 & 10.84 & 0.49\\
		\cmark & \xmark & \cmark & \cmark & \xmark & 16 & 14.78 & 12.31 & 0.53\\
		\cmark & \cmark & \cmark & \cmark & \xmark & 16 & 9.14 & 7.30 & 0.38\\
		\cmark & \cmark & \cmark & \cmark & \cmark & 8 & $\underline{6.06}$ & $\underline{4.73}$ & $\underline{0.14}$ \\
		\cmark & \cmark & \cmark & \cmark & \cmark & 16 & $\textbf{5.88}$ & $\textbf{4.52}$ &$\textbf{0.12}$ \\
		\cmark & \cmark & \cmark & \cmark & \cmark & 32 & 6.21 & 4.85 & 0.15 \\
		\bottomrule
		\end{tabular}
		}
        \caption{Ablation of the gated recurrent memory (Gate), temporal regulation (Temp.), spatial regulation (Spat.), submap-aware memory management (Submap), loop closure (LC), and submap size $W$ on the Apartment dataset\cite{replica}. Enabling both Temp. and Spat. denotes temporal--spatial fused regulation. \cmark/\xmark\hspace{0.3em} indicate module enabled/disabled. The complete model with $W=16$ achieves the best performance.}
		\label{tab:ablation}
		\end{table}

Compared with disabling both cues, which yields reconstruction accuracy and completeness errors of 20.40 cm and 17.39 cm and an RMSE of ATE of 0.70 m, temporal-only regulation improves the results to 13.26 cm, 10.84 cm, and 0.49 m, while spatial-only regulation obtains 14.78 cm, 12.31 cm, and 0.53 m. Combining both cues further reduces the three errors to 9.14 cm, 7.30 cm, and 0.38 m, confirming their complementarity. We also evaluate submap-aware memory management and global SL(4) optimization. Without submap organization, memory contamination becomes more severe in long sequences, leading to lower reconstruction completeness and larger pose errors. Without loop closure or SL(4)-based optimization, the system can still perform local tracking, but global consistency degrades in multi-room and long-trajectory scenarios. With all components enabled, the model achieves the best overall performance, with reconstruction accuracy and completeness errors of 5.88 cm and 4.52 cm, respectively, and an RMSE of ATE value of 0.12 m. These results confirm that local adaptive memory and global submap alignment are complementary: the former improves online local reconstruction, while the latter corrects accumulated drift across submaps.


Finally, we study the influence of sequence length and submap size. Moderate submap sizes provide the best trade-off between local memory stability and global optimization cost. With all components enabled, the full model achieves the best overall balance among pose accuracy, reconstruction completeness, rendering quality, and online efficiency.

\section{Conclusion}
\label{sec:conclusion}

In this paper, we presented a memory-centric 3D foundation model for scalable robotic world modeling. 
Different from existing feed-forward or weakly stateful 3D foundation models, our method maintains an adaptive world memory that supports long-sequence camera localization, dense reconstruction, and Gaussian rendering in an online manner. 
By combining gated recurrent spatial states with test-time temporal--spatial memory regulation, the proposed framework selectively updates, preserves, and forgets memory tokens according to both learned geometric propagation and observation reliability. 
To further support large-scale operation, we organize the world memory into local submaps and introduce loop closure with SL(4)-based submap optimization for global consistency.

Extensive experiments on public benchmarks and self-collected multi-platform robotic datasets demonstrate the effectiveness of the proposed method across video depth estimation, camera pose estimation, dense reconstruction, Gaussian rendering, and long-sequence mapping. 
The results show that adaptive memory regulation improves metric consistency and tracking robustness, while submap-aware memory management enables scalable reconstruction in large indoor and indoor--outdoor environments. 
Moreover, the Gaussian reconstruction head allows the shared memory representation to be decoded into a photorealistic renderable map, extending 3D foundation models from geometric prediction toward persistent robotic world representations.

In future work, we plan to further improve the semantic and dynamic reasoning ability of memory-based 3D foundation models, enabling more robust operation in highly dynamic environments and supporting downstream robotic tasks such as navigation, manipulation, and embodied decision-making.

\bibliographystyle{IEEEtran}
\bibliography{egbib}

@String(CVPR= {IEEE Conf. Comput. Vis. Pattern Recog.})

@String(ECCV= {Eur. Conf. Comput. Vis.})

@String(CVPR  = {CVPR})

@String(ECCV  = {ECCV})

@inproceedings{NeRF,
author = {Mildenhall, Ben and Srinivasan, Pratul P. and Tancik, Matthew and Barron, Jonathan T. and Ramamoorthi, Ravi and Ng, Ren},
title = {NeRF: Representing Scenes as Neural Radiance Fields for View Synthesis},
booktitle = {ECCV},
year = {2020},
}

@InProceedings{niceslam,
    author    = {Zhu, Zihan and Peng, Songyou and Larsson, Viktor and Xu, Weiwei and Bao, Hujun and Cui, Zhaopeng and Oswald, Martin R. and Pollefeys, Marc},
    title     = {NICE-SLAM: Neural Implicit Scalable Encoding for SLAM},
    booktitle = {CVPR},
    month     = {June},
    year      = {2022},
    pages     = {12786-12796}
}

@article{replica,
  title={The Replica dataset: A digital replica of indoor spaces},
  author={Straub, Julian and Whelan, Thomas and Ma, Lingni and Chen, Yufan and Wijmans, Erik and Green, Simon and Engel, Jakob J and Mur-Artal, Raul and Ren, Carl and Verma, Shobhit and others},
  journal={arXiv preprint arXiv:1906.05797},
  year={2019}
}

@ARTICLE{vins,
  author={Qin, Tong and Li, Peiliang and Shen, Shaojie},
  journal={IEEE Transactions on Robotics}, 
  title={VINS-Mono: A Robust and Versatile Monocular Visual-Inertial State Estimator}, 
  year={2018},
  volume={34},
  number={4},
  pages={1004-1020},
  doi={10.1109/TRO.2018.2853729}}

@ARTICLE{orbslam2,
  author={Mur-Artal, Raúl and Tardós, Juan D.},
  journal={IEEE Transactions on Robotics}, 
  title={ORB-SLAM2: An Open-Source SLAM System for Monocular, Stereo, and RGB-D Cameras}, 
  year={2017},
  volume={33},
  number={5},
  pages={1255-1262},
  doi={10.1109/TRO.2017.2705103}}

@article{Bundle-fusion,
author = {Dai, Angela and Nie\ss{}ner, Matthias and Zollh\"{o}fer, Michael and Izadi, Shahram and Theobalt, Christian},
title = {BundleFusion: Real-Time Globally Consistent 3D Reconstruction Using On-the-Fly Surface Reintegration},
year = {2017},
issue_date = {August 2017},
publisher = {Association for Computing Machinery},
address = {New York, NY, USA},
volume = {36},
number = {4},
issn = {0730-0301},
url = {https://doi.org/10.1145/3072959.3054739},
doi = {10.1145/3072959.3054739},
journal = {ACM Trans. Graph.},
month = {jul},
articleno = {76a},
numpages = {18}
}

@ARTICLE{deepfactor,
  author={Czarnowski, Jan and Laidlow, Tristan and Clark, Ronald and Davison, Andrew J.},
  journal={IEEE Robotics and Automation Letters}, 
  title={DeepFactors: Real-Time Probabilistic Dense Monocular SLAM}, 
  year={2020},
  volume={5},
  number={2},
  pages={721-728},
  doi={10.1109/LRA.2020.2965415}}

@InProceedings{neural_rgbd,
    author    = {Azinovi\'c, Dejan and Martin-Brualla, Ricardo and Goldman, Dan B and Nie{\ss}ner, Matthias and Thies, Justus},
    title     = {Neural RGB-D Surface Reconstruction},
    booktitle = {Proceedings of the IEEE/CVF Conference on Computer Vision and Pattern Recognition (CVPR)},
    month     = {June},
    year      = {2022},
    pages     = {6290-6301}
}

@InProceedings{scannet,
author = {Dai, Angela and Chang, Angel X. and Savva, Manolis and Halber, Maciej and Funkhouser, Thomas and Niessner, Matthias},
title = {ScanNet: Richly-Annotated 3D Reconstructions of Indoor Scenes},
booktitle = {Proceedings of the IEEE Conference on Computer Vision and Pattern Recognition (CVPR)},
month = {July},
year = {2017}
}

@INPROCEEDINGS{tum,
  author={Sturm, Jürgen and Engelhard, Nikolas and Endres, Felix and Burgard, Wolfram and Cremers, Daniel},
  booktitle={2012 IEEE/RSJ International Conference on Intelligent Robots and Systems}, 
  title={A benchmark for the evaluation of RGB-D SLAM systems}, 
  year={2012},
  volume={},
  number={},
  pages={573-580},
  doi={10.1109/IROS.2012.6385773}}

@inproceedings{eslam,
  title={Eslam: Efficient dense slam system based on hybrid representation of signed distance fields},
  author={Johari, Mohammad Mahdi and Carta, Camilla and Fleuret, Fran{\c{c}}ois},
  booktitle={Proceedings of the IEEE/CVF Conference on Computer Vision and Pattern Recognition},
  pages={17408--17419},
  year={2023}
}

@inproceedings{plgslam,
  title={Plgslam: Progressive neural scene represenation with local to global bundle adjustment},
  author={Deng, Tianchen and Shen, Guole and Qin, Tong and Wang, Jianyu and Zhao, Wentao and Wang, Jingchuan and Wang, Danwei and Chen, Weidong},
  booktitle={Proceedings of the IEEE/CVF Conference on Computer Vision and Pattern Recognition},
  pages={19657--19666},
  year={2024}
}

@inproceedings{goslam,
  title={Go-slam: Global optimization for consistent 3d instant reconstruction},
  author={Zhang, Youmin and Tosi, Fabio and Mattoccia, Stefano and Poggi, Matteo},
  booktitle={Proceedings of the IEEE/CVF International Conference on Computer Vision},
  pages={3727--3737},
  year={2023}
}

@article{droidslam,
  title={Droid-slam: Deep visual slam for monocular, stereo, and rgb-d cameras},
  author={Teed, Zachary and Deng, Jia},
  journal={Advances in neural information processing systems},
  volume={34},
  pages={16558--16569},
  year={2021}
}

@inproceedings{splatam,
  title={SplaTAM: Splat Track \& Map 3D Gaussians for Dense RGB-D SLAM},
  author={Keetha, Nikhil and Karhade, Jay and Jatavallabhula, Krishna Murthy and Yang, Gengshan and Scherer, Sebastian and Ramanan, Deva and Luiten, Jonathon},
  booktitle={Proceedings of the IEEE/CVF Conference on Computer Vision and Pattern Recognition},
  pages={21357--21366},
  year={2024}
}

@Article{3dgs,
      author       = {Kerbl, Bernhard and Kopanas, Georgios and Leimk{\"u}hler, Thomas and Drettakis, George},
      title        = {3D Gaussian Splatting for Real-Time Radiance Field Rendering},
      journal      = {ACM Transactions on Graphics},
      number       = {4},
      volume       = {42},
      month        = {July},
      year         = {2023},
      url          = {https://repo-sam.inria.fr/fungraph/3d-gaussian-splatting/}
}

@article{orbslam3,
  title={Orb-slam3: An accurate open-source library for visual, visual--inertial, and multimap slam},
  author={Campos, Carlos and Elvira, Richard and Rodr{\'\i}guez, Juan J G{\'o}mez and Montiel, Jos{\'e} MM and Tard{\'o}s, Juan D},
  journal={IEEE Transactions on Robotics},
  volume={37},
  number={6},
  pages={1874--1890},
  year={2021},
  publisher={IEEE}
}

@InProceedings{photoslam,
    author    = {Huang, Huajian and Li, Longwei and Cheng, Hui and Yeung, Sai-Kit},
    title     = {Photo-SLAM: Real-time Simultaneous Localization and Photorealistic Mapping for Monocular Stereo and RGB-D Cameras},
    booktitle = {Proceedings of the IEEE/CVF Conference on Computer Vision and Pattern Recognition (CVPR)},
    month     = {June},
    year      = {2024},
    pages     = {21584-21593}
}

@inproceedings{monogs,
  title={Gaussian splatting slam},
  author={Matsuki, Hidenobu and Murai, Riku and Kelly, Paul HJ and Davison, Andrew J},
  booktitle={Proceedings of the IEEE/CVF Conference on Computer Vision and Pattern Recognition},
  pages={18039--18048},
  year={2024}
}

@article{gsslam,
  title={GS-SLAM: Dense Visual SLAM with 3D Gaussian Splatting},
  author={Yan, Chi and Qu, Delin and Wang, Dong and Xu, Dan and Wang, Zhigang and Zhao, Bin and Li, Xuelong},
  journal={arXiv preprint arXiv:2311.11700},
  year={2023}
}

@article{dpvo,
  title={Deep patch visual odometry},
  author={Teed, Zachary and Lipson, Lahav and Deng, Jia},
  journal={Advances in Neural Information Processing Systems},
  volume={36},
  year={2024}
}

@article{kintinuous,
  title={Kintinuous: Spatially extended kinectfusion},
  author={Whelan, Thomas and Kaess, Michael and Fallon, Maurice and Johannsson, Hordur and Leonard, John and McDonald, John},
  year={2012}
}

@article{elasticfusion,
  title={ElasticFusion: Real-time dense SLAM and light source estimation},
  author={Whelan, Thomas and Salas-Moreno, Renato F and Glocker, Ben and Davison, Andrew J and Leutenegger, Stefan},
  journal={The International Journal of Robotics Research},
  volume={35},
  number={14},
  pages={1697--1716},
  year={2016},
  publisher={SAGE Publications Sage UK: London, England}
}

@inproceedings{dust3r,
  title={Dust3r: Geometric 3d vision made easy},
  author={Wang, Shuzhe and Leroy, Vincent and Cabon, Yohann and Chidlovskii, Boris and Revaud, Jerome},
  booktitle={Proceedings of the IEEE/CVF Conference on Computer Vision and Pattern Recognition},
  pages={20697--20709},
  year={2024}
}

@inproceedings{vggt,
  title={Vggt: Visual geometry grounded transformer},
  author={Wang, Jianyuan and Chen, Minghao and Karaev, Nikita and Vedaldi, Andrea and Rupprecht, Christian and Novotny, David},
  booktitle={Proceedings of the Computer Vision and Pattern Recognition Conference},
  pages={5294--5306},
  year={2025}
}

@inproceedings{mneslam,
  title={MNE-SLAM: Multi-Agent Neural SLAM for Mobile Robots},
  author={Deng, Tianchen and Shen, Guole and Xun, Chen and Yuan, Shenghai and Jin, Tongxin and Shen, Hongming and Wang, Yanbo and Wang, Jingchuan and Wang, Hesheng and Wang, Danwei and others},
  booktitle={Proceedings of the Computer Vision and Pattern Recognition Conference},
  pages={1485--1494},
  year={2025}
}

@inproceedings{mast3r,
  title={Grounding image matching in 3d with mast3r},
  author={Leroy, Vincent and Cabon, Yohann and Revaud, J{\'e}r{\^o}me},
  booktitle={European Conference on Computer Vision},
  pages={71--91},
  year={2024},
  organization={Springer}
}

@article{cut3r,
  title={Continuous 3D Perception Model with Persistent State},
  author={Wang, Qianqian and Zhang, Yifei and Holynski, Aleksander and Efros, Alexei A and Kanazawa, Angjoo},
  journal={arXiv preprint arXiv:2501.12387},
  year={2025}
}

@article{spann3r,
  title={3d reconstruction with spatial memory},
  author={Wang, Hengyi and Agapito, Lourdes},
  journal={arXiv preprint arXiv:2408.16061},
  year={2024}
}

@article{mast3rslam,
  title={MASt3R-SLAM: Real-Time Dense SLAM with 3D Reconstruction Priors},
  author={Murai, Riku and Dexheimer, Eric and Davison, Andrew J},
  journal={arXiv preprint arXiv:2412.12392},
  year={2024}
}

@inproceedings{7scenes,
  title={Scene coordinate regression forests for camera relocalization in RGB-D images},
  author={Shotton, Jamie and Glocker, Ben and Zach, Christopher and Izadi, Shahram and Criminisi, Antonio and Fitzgibbon, Andrew},
  booktitle={Proceedings of the IEEE conference on computer vision and pattern recognition},
  pages={2930--2937},
  year={2013}
}

@inproceedings{nicerslam,
  title={Nicer-slam: Neural implicit scene encoding for rgb slam},
  author={Zhu, Zihan and Peng, Songyou and Larsson, Viktor and Cui, Zhaopeng and Oswald, Martin R and Geiger, Andreas and Pollefeys, Marc},
  booktitle={2024 International Conference on 3D Vision (3DV)},
  pages={42--52},
  year={2024},
  organization={IEEE}
}

@article{deng2025best3dscenerepresentation,
      title={What Is The Best 3D Scene Representation for Robotics? From Geometric to Foundation Models}, 
      author={Tianchen Deng and Yue Pan and Shenghai Yuan and Dong Li and Chen Wang and Mingrui Li and Long Chen and Lihua Xie and Danwei Wang and Jingchuan Wang and Javier Civera and Hesheng Wang and Weidong Chen},
      year={2025},
      journal={arXiv preprint arXiv:2512.03422}, 
}

@article{slam3r,
  title={SLAM3R: Real-Time Dense Scene Reconstruction from Monocular RGB Videos},
  author={Liu, Yuzheng and Dong, Siyan and Wang, Shuzhe and Yin, Yingda and Yang, Yanchao and Fan, Qingnan and Chen, Baoquan},
  journal={arXiv preprint arXiv:2412.09401},
  year={2024}
}

@article{wang2025pi,
  title={$\pi^3$: Permutation-Equivariant Visual Geometry Learning},
  author={Wang, Yifan and Zhou, Jianjun and Zhu, Haoyi and Chang, Wenzheng and Zhou, Yang and Li, Zizun and Chen, Junyi and Pang, Jiangmiao and Shen, Chunhua and He, Tong},
  journal={arXiv preprint arXiv:2507.13347},
  year={2025}
}

@article{shen2025grs,
  title={Grs-slam3r: Real-time dense slam with gated recurrent state},
  author={Shen, Guole and Deng, Tianchen and Wang, Yanbo and Chen, Yongtao and Shen, Yilin and Liu, Jiuming and Wang, Jingchuan},
  journal={arXiv preprint arXiv:2509.23737},
  year={2025}
}

@article{wu2026point3r,
  title={Point3r: Streaming 3d reconstruction with explicit spatial pointer memory},
  author={Wu, Yuqi and Zheng, Wenzhao and Zhou, Jie and Lu, Jiwen},
  journal={Advances in Neural Information Processing Systems},
  volume={38},
  pages={69675--69699},
  year={2026}
}

@inproceedings{chenttt3r,
  title={TTT3R: 3D Reconstruction as Test-Time Training},
  author={Chen, Xingyu and Chen, Yue and Xiu, Yuliang and Geiger, Andreas and Chen, Anpei},
  booktitle={The Fourteenth International Conference on Learning Representations}
}

@article{lan2025stream3r,
  title={Stream3r: Scalable sequential 3d reconstruction with causal transformer},
  author={Lan, Yushi and Luo, Yihang and Hong, Fangzhou and Zhou, Shangchen and Chen, Honghua and Lyu, Zhaoyang and Yang, Shuai and Dai, Bo and Loy, Chen Change and Pan, Xingang},
  journal={arXiv preprint arXiv:2508.10893},
  year={2025}
}

@article{zheng2026ttsa3r,
  title={TTSA3R: Training-Free Temporal-Spatial Adaptive Persistent State for Streaming 3D Reconstruction},
  author={Zheng, Zhijie and Xiang, Xinhao and Zhang, Jiawei},
  journal={arXiv preprint arXiv:2601.22615},
  year={2026}
}

@article{zhuo2025streaming,
  title={Streaming 4d visual geometry transformer},
  author={Zhuo, Dong and Zheng, Wenzhao and Guo, Jiahe and Wu, Yuqi and Zhou, Jie and Lu, Jiwen},
  journal={arXiv preprint arXiv:2507.11539},
  year={2025}
}

@article{sun2024learning,
  title={Learning to (learn at test time): Rnns with expressive hidden states},
  author={Sun, Yu and Li, Xinhao and Dalal, Karan and Xu, Jiarui and Vikram, Arjun and Zhang, Genghan and Dubois, Yann and Chen, Xinlei and Wang, Xiaolong and Koyejo, Sanmi and others},
  journal={arXiv preprint arXiv:2407.04620},
  year={2024}
}

@article{zhang2025zisvfm,
  title={ZISVFM: Zero-shot object instance segmentation in indoor robotic environments with vision foundation models},
  author={Zhang, Ying and Yin, Maoliang and Bi, Wenfu and Yan, Haibao and Bian, Shaohan and Zhang, Cui-Hua and Hua, Changchun},
  journal={IEEE Transactions on Robotics},
  volume={41},
  pages={1568--1580},
  year={2025},
  publisher={IEEE}
}

@article{talak2023certifiable,
  title={Certifiable object pose estimation: Foundations, learning models, and self-training},
  author={Talak, Rajat and Peng, Lisa R and Carlone, Luca},
  journal={IEEE Transactions on Robotics},
  volume={39},
  number={4},
  pages={2805--2824},
  year={2023},
  publisher={IEEE}
}

@article{shen2026unilgl,
  title={UniLGL: learning uniform place recognition for FOV-limited/Panoramic LiDAR global localization},
  author={Shen, Hongming and Chen, Xun and Hui, Yulin and Wu, Zhenyu and Wang, Wei and Lyu, Qiyang and Deng, Tianchen and Wang, Danwei},
  journal={IEEE Transactions on Robotics},
  year={2026},
  publisher={IEEE}
}

@article{yuan2026infinitevggt,
  title={InfiniteVGGT: Visual Geometry Grounded Transformer for Endless Streams},
  author={Yuan, Shuai and Yang, Yantai and Yang, Xiaotian and Zhang, Xupeng and Zhao, Zhonghao and Zhang, Lingming and Zhang, Zhipeng},
  journal={arXiv preprint arXiv:2601.02281},
  year={2026}
}

@article{deng2024compact,
  title={Compact 3d gaussian splatting for dense visual slam},
  author={Deng, Tianchen and Chen, Yaohui and Zhang, Leyan and Yang, Jianfei and Yuan, Shenghai and Liu, Jiuming and Wang, Danwei and Wang, Hesheng and Chen, Weidong},
  journal={arXiv preprint arXiv:2403.11247},
  year={2024}
}

@inproceedings{xie2026scal3r,
  title={Scal3r: Scalable test-time training for large-scale 3d reconstruction},
  author={Xie, Tao and Yang, Peishan and Jin, Yudong and Cai, Yingfeng and Yin, Wei and Ren, Weiqiang and Zhang, Qian and Hua, Wei and Peng, Sida and Guo, Xiaoyang and others},
  booktitle={Proceedings of the IEEE/CVF Conference on Computer Vision and Pattern Recognition},
  pages={21760--21771},
  year={2026}
}

@article{jin2026zipmap,
  title={ZipMap: Linear-Time Stateful 3D Reconstruction via Test-Time Training},
  author={Jin, Haian and Wu, Rundi and Zhang, Tianyuan and Gao, Ruiqi and Barron, Jonathan T and Snavely, Noah and Holynski, Aleksander},
  journal={arXiv preprint arXiv:2603.04385},
  year={2026}
}

@article{zhang2026loger,
  title={Loger: Long-context geometric reconstruction with hybrid memory},
  author={Zhang, Junyi and Herrmann, Charles and Hur, Junhwa and Sun, Chen and Yang, Ming-Hsuan and Cole, Forrester and Darrell, Trevor and Sun, Deqing},
  journal={arXiv preprint arXiv:2603.03269},
  year={2026}
}

@article{liu2026mem3r,
  title={Mem3R: Streaming 3D Reconstruction with Hybrid Memory via Test-Time Training},
  author={Liu, Changkun and Yang, Jiezhi and Li, Zeman and Deng, Yuan and Guo, Jiancong and Ballan, Luca},
  journal={arXiv preprint arXiv:2604.07279},
  year={2026}
}

@article{deng2025vggt,
  title={VGGT-Long: Chunk it, Loop it, Align it--Pushing VGGT's Limits on Kilometer-scale Long RGB Sequences},
  author={Deng, Kai and Ti, Zexin and Xu, Jiawei and Yang, Jian and Xie, Jin},
  journal={arXiv preprint arXiv:2507.16443},
  year={2025}
}

@article{Lact,
  title={Test-time training done right},
  author={Zhang, Tianyuan and Bi, Sai and Hong, Yicong and Zhang, Kai and Luan, Fujun and Yang, Songlin and Sunkavalli, Kalyan and Freeman, William T and Tan, Hao},
  journal={arXiv preprint arXiv:2505.23884},
  year={2025}
}

@article{wang2025amb3r,
  title={AMB3R: Accurate Feed-forward Metric-scale 3D Reconstruction with Backend},
  author={Wang, Hengyi and Agapito, Lourdes},
  journal={arXiv preprint arXiv:2511.20343},
  year={2025}
}

@article{elflein2026vgg,
  title={VGG-T $^3$: Offline Feed-Forward 3D Reconstruction at Scale},
  author={Elflein, Sven and Li, Ruilong and Agostinho, S{\'e}rgio and Gojcic, Zan and Leal-Taix{\'e}, Laura and Zhou, Qunjie and Osep, Aljosa},
  journal={arXiv preprint arXiv:2602.23361},
  year={2026}
}

@inproceedings{fang2026incvggt,
  title={IncVGGT: Incremental VGGT for Memory-Bounded Long-Range 3D Reconstruction},
  author={Fang, Keyu and Zhou, Changchun and Fu, Yuzhe and Li, Hai Helen and Chen, Yiran},
  booktitle={The Fourteenth International Conference on Learning Representations},
  year={2026}
}

@article{deng2026mamba,
  title={Mamba-VGGT: Persistent Long-Sequence Video Geometry Grounded Transformer via External Sliding Window Mamba Memory},
  author={Deng, Tianchen and Xiong, Zhenxiang and Wang, Nailin and Wang, Fangjinhua and Liu, Jiuming and Yang, Jianfei and Wang, Hesheng},
  journal={arXiv preprint arXiv:2605.17478},
  year={2026}
}

@inproceedings{katharopoulos2020transformers,
  title={Transformers are rnns: Fast autoregressive transformers with linear attention},
  author={Katharopoulos, Angelos and Vyas, Apoorv and Pappas, Nikolaos and Fleuret, Fran{\c{c}}ois},
  booktitle={International conference on machine learning},
  pages={5156--5165},
  year={2020},
  organization={PMLR}
}

@article{gu2023mamba,
  title={Mamba: Linear-time sequence modeling with selective state spaces},
  author={Gu, Albert and Dao, Tri},
  journal={arXiv preprint arXiv:2312.00752},
  year={2023}
}

@inproceedings{schlag2021linear,
  title={Linear transformers are secretly fast weight programmers},
  author={Schlag, Imanol and Irie, Kazuki and Schmidhuber, J{\"u}rgen},
  booktitle={International conference on machine learning},
  pages={9355--9366},
  year={2021},
  organization={PMLR}
}

@article{maggio2026vggt,
  title={Vggt-slam: Dense rgb slam optimized on the sl (4) manifold},
  author={Maggio, Dominic and Lim, Hyungtae and Carlone, Luca},
  journal={Advances in Neural Information Processing Systems},
  volume={38},
  pages={129839--129867},
  year={2026}
}

@article{GRU,
  title={Empirical evaluation of gated recurrent neural networks on sequence modeling},
  author={Chung, Junyoung and Gulcehre, Caglar and Cho, KyungHyun and Bengio, Yoshua},
  journal={arXiv preprint arXiv:1412.3555},
  year={2014}
}

@article{GRU2,
  title={On the properties of neural machine translation: Encoder-decoder approaches},
  author={Cho, Kyunghyun and Van Merri{\"e}nboer, Bart and Bahdanau, Dzmitry and Bengio, Yoshua},
  journal={arXiv preprint arXiv:1409.1259},
  year={2014}
}

@article{unipr-3d,
      title={UniPR-3D: Towards Universal Visual Place Recognition with Visual Geometry Grounded Transformer}, 
      author={Tianchen Deng and Xun Chen and Ziming Li and Hongming Shen and Danwei Wang and Javier Civera and Hesheng Wang},
      year={2025},
     journal={arXiv preprint 2512.21078}, 
}

@inproceedings{sintel,
title = {A naturalistic open source movie for optical flow evaluation},
author = {Butler, D. J. and Wulff, J. and Stanley, G. B. and Black, M. J.},
booktitle = {European Conf. on Computer Vision (ECCV)},
editor = {{A. Fitzgibbon et al. (Eds.)}},
publisher = {Springer-Verlag},
series = {Part IV, LNCS 7577},
month = oct,
pages = {611--625},
year = {2012}
}

@InProceedings{bonn,
author = {E. Palazzolo and J. Behley and P. Lottes and P. Gigu\`ere and C. Stachniss},
title = {{ReFusion: 3D Reconstruction in Dynamic Environments for RGB-D Cameras Exploiting Residuals}},
booktitle = iros,
year = {2019},
url = {https://www.ipb.uni-bonn.de/pdfs/palazzolo2019iros.pdf},
codeurl = {https://github.com/PRBonn/refusion},
videourl = {https://youtu.be/1P9ZfIS5-p4},
}

@article{kitti,
  author = {Andreas Geiger and Philip Lenz and Christoph Stiller and Raquel Urtasun},
  title = {Vision meets Robotics: The KITTI Dataset},
  journal = {International Journal of Robotics Research (IJRR)},
  year = {2013}
}

@inproceedings{zhang2025monst3r,
  title={Monst3r: A simple approach for estimating geometry in the presence of motion},
  author={Zhang, Junyi and Herrmann, Charles and Hur, Junhwa and Jampani, Varun and Cole, Forrester and Sun, Deqing and Yang, Ming-Hsuan and others},
  booktitle={International Conference on Learning Representations},
  volume={2025},
  pages={82863--82886},
  year={2025}
}

@inproceedings{chen2025easi3r,
  title={Easi3r: Estimating disentangled motion from dust3r without training},
  author={Chen, Xingyu and Chen, Yue and Xiu, Yuliang and Geiger, Andreas and Chen, Anpei},
  booktitle={Proceedings of the IEEE/CVF International Conference on Computer Vision},
  pages={9158--9168},
  year={2025}
}

@article{shen2025mut3r,
  title={MUT3R: Motion-aware Updating Transformer for Dynamic 3D Reconstruction},
  author={Shen, Guole and Deng, Tianchen and Qin, Xingrui and Wang, Nailin and Wang, Jianyu and Wang, Yanbo and Chen, Yongtao and Wang, Hesheng and Wang, Jingchuan},
  journal={arXiv preprint arXiv:2512.03939},
  year={2025}
}

@inproceedings{kopf2021robust,
  title={Robust consistent video depth estimation},
  author={Kopf, Johannes and Rong, Xuejian and Huang, Jia-Bin},
  booktitle={Proceedings of the IEEE/CVF Conference on Computer Vision and Pattern Recognition},
  pages={1611--1621},
  year={2021}
}

@inproceedings{zhang2022structure,
  title={Structure and motion from casual videos},
  author={Zhang, Zhoutong and Cole, Forrester and Li, Zhengqi and Rubinstein, Michael and Snavely, Noah and Freeman, William T},
  booktitle={European Conference on Computer Vision},
  pages={20--37},
  year={2022},
  organization={Springer}
}

@article{teed2018deepv2d,
  title={Deepv2d: Video to depth with differentiable structure from motion},
  author={Teed, Zachary and Deng, Jia},
  journal={arXiv preprint arXiv:1812.04605},
  year={2018}
}

@inproceedings{zhang2026vista,
  title={Vista-slam: Visual slam with symmetric two-view association},
  author={Zhang, Ganlin and Qian, Shenhan and Wang, Xi and Cremers, Daniel},
  booktitle={2026 International Conference on 3D Vision (3DV)},
  pages={396--406},
  year={2026},
  organization={IEEE}
}

@inproceedings{gao2018ldso,
  title={LDSO: Direct sparse odometry with loop closure},
  author={Gao, Xiang and Wang, Rui and Demmel, Nikolaus and Cremers, Daniel},
  booktitle={2018 IEEE/RSJ International Conference on Intelligent Robots and Systems (IROS)},
  pages={2198--2204},
  year={2018},
  organization={IEEE}
}

@inproceedings{lipson2024deep,
  title={Deep patch visual slam},
  author={Lipson, Lahav and Teed, Zachary and Deng, Jia},
  booktitle={European Conference on Computer Vision},
  pages={424--440},
  year={2024},
  organization={Springer}
}

@inproceedings{yang2025fast3r,
  title={Fast3r: Towards 3d reconstruction of 1000+ images in one forward pass},
  author={Yang, Jianing and Sax, Alexander and Liang, Kevin J and Henaff, Mikael and Tang, Hao and Cao, Ang and Chai, Joyce and Meier, Franziska and Feiszli, Matt},
  booktitle={Proceedings of the Computer Vision and Pattern Recognition Conference},
  pages={21924--21935},
  year={2025}
}

@article{li2023dense,
  title={Dense rgb slam with neural implicit maps},
  author={Li, Heng and Gu, Xiaodong and Yuan, Weihao and Yang, Luwei and Dong, Zilong and Tan, Ping},
  journal={arXiv preprint arXiv:2301.08930},
  year={2023}
}

@inproceedings{zhao2025pseudo,
  title={Pseudo Depth Meets Gaussian: A Feed-forward RGB SLAM Baseline},
  author={Zhao, Linqing and Xu, Xiuwei and Wang, Yirui and Wang, Hao and Zheng, Wenzhao and Tang, Yansong and Yan, Haibin and Lu, Jiwen},
  booktitle={2025 IEEE/RSJ International Conference on Intelligent Robots and Systems (IROS)},
  pages={8142--8149},
  year={2025},
  organization={IEEE}
}

@inproceedings{cheng2025outdoor,
  title={Outdoor monocular slam with global scale-consistent 3d gaussian pointmaps},
  author={Cheng, Chong and Yu, Sicheng and Wang, Zijian and Zhou, Yifan and Wang, Hao},
  booktitle={Proceedings of the IEEE/CVF International Conference on Computer Vision},
  pages={26035--26044},
  year={2025}
}

@article{zhang2026flash,
  title={Flash-Mono: Feed-Forward Accelerated Gaussian Splatting Monocular SLAM},
  author={Zhang, Zicheng and Wu, Ke and Meng, Xiangting and Liu, Keyu and Zhao, Jieru and Ding, Wenchao},
  journal={arXiv preprint arXiv:2604.03092},
  year={2026}
}

@inproceedings{Geiger2012CVPR,
  author = {Andreas Geiger and Philip Lenz and Raquel Urtasun},
  title = {Are we ready for Autonomous Driving? The KITTI Vision Benchmark Suite},
  booktitle = {Conference on Computer Vision and Pattern Recognition (CVPR)},
  year = {2012}
}

\end{document}